\documentclass{article}

\usepackage{microtype}
\usepackage{graphicx}
\usepackage{subcaption}
\usepackage{booktabs} 
\usepackage{fontawesome5}

\usepackage{hyperref}

\newcommand{\AlgPhase}[1]{\vspace{3pt}\Statex\textbf{\textsc{#1}}}

\usepackage{algpseudocode}
\usepackage{algorithm}

\usepackage[preprint]{icml2026}

\usepackage{amsmath}
\usepackage{amssymb}
\usepackage{mathtools}
\usepackage{amsthm}
\usepackage{xspace}
\usepackage{booktabs}       
\usepackage{amsfonts}       
\usepackage{nicefrac}       
\usepackage{microtype}      
\usepackage{xcolor}         
\usepackage{multirow}
\usepackage{amsmath}
\usepackage{booktabs}
\usepackage{graphicx}
\usepackage{caption}
\usepackage{array}
\usepackage{marvosym}
\usepackage{listings}

\usepackage{enumitem}
\usepackage[table]{xcolor}
\lstdefinelanguage{json}{
    basicstyle=\ttfamily\scriptsize,
    numbers=none,
    showstringspaces=false,
    breaklines=true,
    frame=lines,
    backgroundcolor=\color{white},
    stringstyle=\color{black},
    keywordstyle=\bfseries\color{black},
    commentstyle=\itshape\color{gray},
    morestring=[b]",
    morestring=[d]'
}

\newcolumntype{C}[1]{>{\centering\let\newline\\\arraybackslash\hspace{0pt}}m{#1}}

\usepackage[capitalize,noabbrev]{cleveref}

\theoremstyle{plain}

\theoremstyle{definition}

\theoremstyle{remark}

\newcommand{\ours}{UI-Mem\xspace}

\usepackage[textsize=tiny]{todonotes}

\icmltitlerunning{UI-Mem: Self-Evolving Experience Memory for Online Reinforcement Learning in Mobile GUI Agents}

\begin{document}

\twocolumn[
  \icmltitle{UI-Mem: Self-Evolving Experience Memory for \\
  Online Reinforcement Learning in Mobile GUI Agents}



  \icmlsetsymbol{equal}{*}
  \icmlsetsymbol{co}{\dag}


  \begin{icmlauthorlist}
    \icmlauthor{Han Xiao}{cuhk,equal}
    \icmlauthor{Guozhi Wang}{vivo,equal}
    \icmlauthor{Hao Wang}{vivo,equal}
    \icmlauthor{Shilong Liu}{princeton}
    \icmlauthor{Yuxiang Chai}{cuhk}
    \icmlauthor{Yue Pan}{vivo}
    \icmlauthor{Yufeng Zhou}{vivo}
    \icmlauthor{Xiaoxin Chen}{vivo}
    \icmlauthor{Yafei Wen}{vivo}
    \icmlauthor{Hongsheng Li}{cuhk,sz,sh,co}
  \end{icmlauthorlist}

  \icmlaffiliation{cuhk}{Multimedia Laboratory (MMLab), The Chinese University
of Hong Kong}
  \icmlaffiliation{sz}{Shenzhen Loop Area Institute}
  \icmlaffiliation{sh}{Shanghai AI Lab}
  \icmlaffiliation{vivo}{vivo AI Lab}
  \icmlaffiliation{princeton}{Princeton University}

  \icmlcorrespondingauthor{Han Xiao}{1155229123@link.cuhk.edu.edu}
  \icmlcorrespondingauthor{Hongsheng Li}{hsli@ee.cuhk.edu.hk}


    \begin{center}
    \vspace{-0.2em}
    \href{https://ui-mem.github.io/}{\faGlobe\ Project Page: https://ui-mem.github.io/}
  \end{center}

  \vskip 0.3in
]



\printAffiliationsAndNotice{}  

\begin{abstract}
   
Online Reinforcement Learning (RL) offers a promising paradigm for enhancing GUI agents through direct environment interaction. However, its effectiveness is severely hindered by inefficient credit assignment in long-horizon tasks and repetitive errors across tasks due to the lack of experience transfer. To address these challenges, we propose UI-Mem, a novel framework that enhances GUI online RL with a Hierarchical Experience Memory. Unlike traditional replay buffers, our memory accumulates structured knowledge, including high-level workflows, subtask skills, and failure patterns. These experiences are stored as parameterized templates that enable cross-task and cross-application transfer. To effectively integrate memory guidance into online RL, we introduce Stratified Group Sampling, which injects varying levels of guidance across trajectories within each rollout group to maintain outcome diversity, driving the unguided policy toward internalizing guided behaviors. Furthermore, a Self-Evolving Loop continuously abstracts novel strategies and errors to keep the memory aligned with the agent's evolving policy. Experiments on online GUI benchmarks demonstrate that UI-Mem significantly outperforms traditional RL baselines and static reuse strategies, with strong generalization to unseen applications.

\end{abstract}
\section{Introduction}
\begin{figure}[t]
  \centering
  \includegraphics[width=0.95\linewidth]{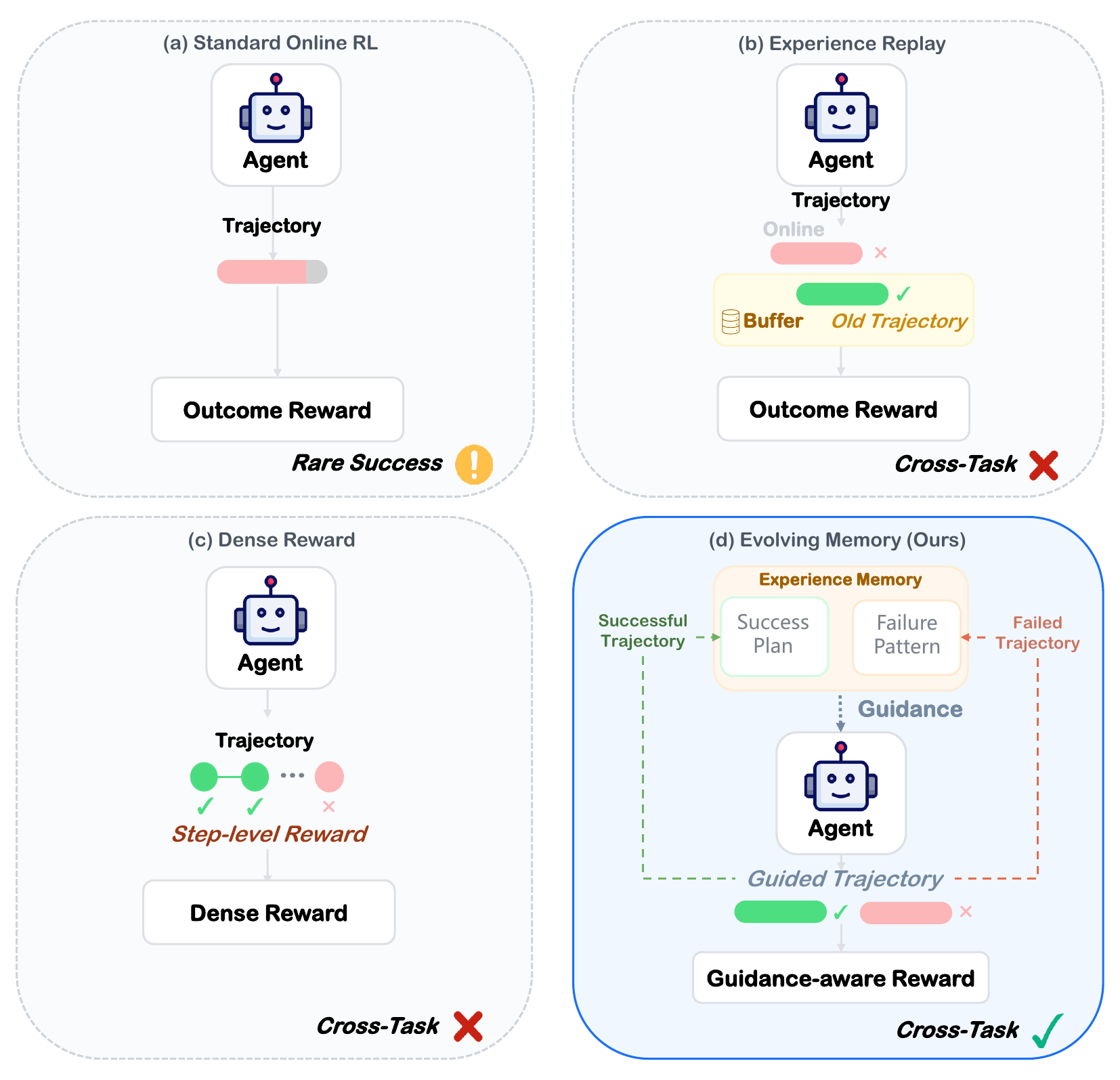}
  \caption{\textbf{Comparison of RL paradigms for GUI agents.} (a) Standard Online RL suffers from sparse rewards. (b) Experience Replay and (c) Dense Reward address sample efficiency and credit assignment respectively, but both lack mechanisms for Cross-Task Transfer. (d) Our Framework introduces an Evolving Memory that provides hierarchical guidance for exploration and continuously updates itself by abstracting successful plans and failure patterns from new trajectories, enabling cross-task knowledge transfer.}
  \label{fig:intro_comparison}
  \vspace{-5mm}
\end{figure}

Recent advances in Multi-modal Large Language Models (MLLMs)~\cite{liu2023visual,Bai2023QwenVLAF,achiam2023gpt} have enabled their application as autonomous GUI agents~\cite{deng2023mind2web,zhang2025appagent,qin2025ui}, capable of interpreting screenshots and generating sequential actions to accomplish user goals. To further enhance the decision-making capabilities of these agents, Reinforcement Learning (RL)~\cite{liu2025infigui, lu2025ui} has emerged as a promising paradigm. In particular, Online RL methods~\cite{bai2024digirl, xu2025mobilerl} enable agents to learn optimal policies through direct environment interaction, generating diverse rollouts and receiving outcome-based feedback via algorithms like Group Relative Policy Optimization (GRPO)~\cite{shao2024deepseekmath}.

However, the effectiveness of standard online RL in GUI environments is severely impeded by the combination of long-horizon tasks and extremely sparse reward signals. In such settings, agents are often trapped in a cycle of blind trial-and-error, facing two primary bottlenecks:
\textbf{(1) Inefficient credit assignment}: even if a trajectory contains several correct intermediate steps, such as navigating to the correct page, a single final error results in negative feedback, preventing the agent from reinforcing its partial successes.
\textbf{(2) Repetitive errors across tasks}: the agent often encounters similar failure modes (e.g., handling a confirmation popup) across different high-level tasks. Lacking a mechanism to store and transfer this experience, the agent must rediscover errors from scratch in every new task.

Existing approaches only partially address these bottlenecks. Data reuse methods, such as Experience Replay~\cite{xu2025mobilerl, lu2025arpo}, stabilize training by injecting past successful trajectories when the on-policy batch contains mostly failures. However, they fail to generate meaningful trajectories when facing novel tasks, especially early in training when success is rare. Reward refinement methods~\cite{wang2023mathshepherd,feng2025group} introduce step-level dense rewards to provide fine-grained supervision. While they identify which step was correct within the current rollout, they do not enable transfer of reusable experience across different tasks and applications. 

Our key insight is that RL for GUI agents should not merely learn from raw trajectories or rely on dense supervision; it should \textit{accumulate, reuse, and evolve} experience that transfers across tasks.
Motivated by this, we propose \ours (\cref{fig:intro_comparison}), a novel framework that enhances GUI online RL with a \textbf{Hierarchical Experience Memory}. 
Unlike traditional replay buffers that store raw trajectories, our memory accumulates structured knowledge: high-level workflows for planning, subtask skills for execution, and failure patterns to prevent repetitive errors. To efficiently utilize the memory, we store this structured knowledge as parameterized templates.
When given a novel task, the agent retrieves relevant abstract templates (e.g., ``Send email to \texttt{\{\{recipient}\}\}'') and instantiates them with current task details to form a concrete plan. 
This design addresses both bottlenecks: hierarchical decomposition allows the agent to receive credit for completing intermediate subtasks even when the full task fails, while template abstraction enables skill reuse across different applications.

However, integrating such memory into online RL introduces new challenges.
Simply prepending retrieved plans to every rollout prompt risks the agent learning to follow external instructions rather than internalizing skills. 
Moreover, if strong guidance causes all trajectories in a rollout group to succeed, the advantage variance drops to zero, eliminating the learning signal.
Therefore, we introduce a \textbf{Stratified Group Sampling} mechanism during the memory-guided exploration process.
Instead of uniform prompting, we construct a diverse batch for GRPO by injecting varying levels of guidance into the prompts of different trajectories within the same group. Some trajectories receive full hierarchical plans to improve success rates, while others receive partial or no guidance to encourage independent exploration. This creates within-group outcome diversity essential for advantage estimation, driving the unguided policy to bridge the performance gap towards the guided trajectories. We employ a dynamic curriculum to progressively reduce guidance as task success rates improve. 
Finally, as the agent's policy improves, it discovers novel strategies and encounters new failure modes absent from the existing memory. 
To ensure that the memory accommodates evolving task distributions, we introduce a \textbf{Self-Evolving Loop}.
We abstract successful trajectories into new plans and extract failure patterns from errors, dynamically updating the memory pool to co-evolve with the agent's policy.
Extensive evaluations on online GUI benchmarks demonstrate that our method significantly outperforms traditional RL baselines and static data reuse strategies across varying model scales. Furthermore, the generalizability to unseen applications highlights the effectiveness of our abstraction and retrieval mechanisms.

\section{Related Work}
\paragraph{Multi-modal GUI Agents.} Recent advances in Large Language Models (LLMs)~\cite{OpenAI2023ChatGPT,OpenAI2023GPT4TR} and Multi-modal LLMs (MLLMs)~\cite{liu2023visual,Bai2023QwenVLAF} have significantly empowered agents to perceive and interact with Graphical User Interfaces (GUIs). 
Early efforts~\cite{wang2024mobile2, zhang2025appagent} build agent frameworks upon commercial MLLMs and demonstrate promising results in mobile and desktop environments.
Recent research has shifted towards fine-tuning open-source MLLMs. For instance, OS-Atlas~\cite{wu2024atlas} and Aguvis~\cite{xu2024aguvis} focus on enhancing UI grounding and propose a unified action space for cross-platform generalization. 
UI-TARS~\cite{qin2025ui} further scales up instruction tuning with massive datasets covering element-wise grounding and System-2 reasoning. 
UI-Genie~\cite{xiao2025ui} introduces a self-evolving framework for synthesizing high-quality trajectories.
Despite these successes in Supervised Fine-Tuning (SFT), these agents often struggle with multi-step reasoning and generalization to novel tasks outside their training distribution, which requires further optimization through environment interaction.
\begin{figure*}[t]
  \centering
  \includegraphics[width=0.98\textwidth]{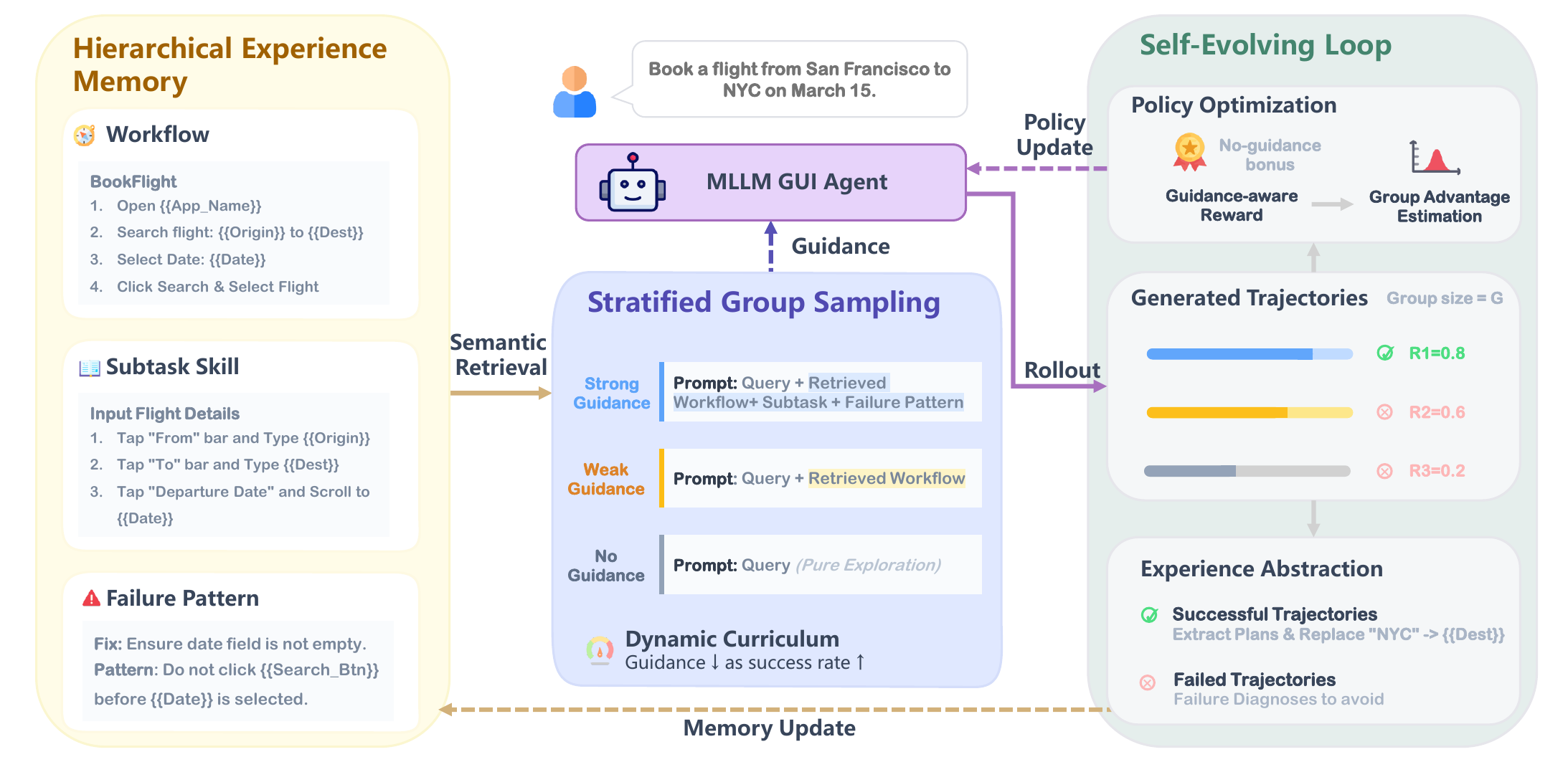}
  \caption{\textbf{Overview of the proposed \ours framework.} Given a task instruction, the agent retrieves hierarchical experience including Workflows, Subtask Skills, and Failure Patterns. We employ \textbf{Stratified Group Sampling} to generate a group of trajectories under varying levels of guidance (Strong, Weak, and No Guidance), enabling effective advantage estimation for \textbf{Policy Optimization}. Finally, a \textbf{Self-Evolving Loop} extracts abstract plans from successful trajectories and diagnoses from failures to update the memory, facilitating continuous refinement and cross-task transfer.}
  \label{fig:pipeline}
  \vspace{-4mm}
\end{figure*}
\paragraph{Reinforcement Learning for Language and Agents.}
Reinforcement Learning has emerged as a powerful paradigm for aligning language models with human preferences~\cite{christiano2017deep} and improving reasoning capabilities~\cite{schulman2017proximal, rafailov2023direct} beyond SFT. 
Recent advancements such as Group Relative Policy Optimization (GRPO)~\cite{shao2024deepseekmath} offer efficient optimization by leveraging group-based advantage estimation without a critic network.
In GUI domains, several works~\cite{lu2025ui,liu2025infigui} have explored GRPO-based optimization with rule-based rewards in an offline manner.
However, applying online RL to GUIs remains challenging due to extremely sparse rewards.
To mitigate this, recent works improve reward design with step-level feedback~\cite{wang2023mathshepherd, feng2025group}, while others rely on Experience Replay to stabilize training~\cite{bai2024digirl, xu2025mobilerl,lu2025arpo}.
Despite their effectiveness, these approaches lack mechanisms to transfer reusable experience across tasks, failing to address the fundamental bottleneck of generating high-quality trajectories in the first place.
By contrast, our \ours constructs a hierarchical, evolving experience memory that actively guides online sampling, significantly improving both trajectory quality and training efficiency.

\section{Method}

\subsection{Overview}

We formulate the GUI interaction as an MDP and adopt GRPO~\cite{shao2024deepseekmath} as our base optimizer. Formal definitions are provided in Appendix (\cref{app:pre}). However, standard GRPO struggles in the context of online GUI RL. 
Due to the vast exploration spaces and long horizons, the sampled group often consists entirely of failures. This results in zero variance in the advantage term, providing no gradient for optimization. 
Furthermore, even when partial success is achieved in some trajectories, the agent lacks a mechanism to retain this experience, forcing it to rediscover solutions for similar tasks from scratch.

To address these challenges, we propose \ours, a novel framework that integrates structured experience memory into the online RL loop. As illustrated in \cref{fig:pipeline}, \ours consists of three core components: \textbf{(1) Hierarchical Experience Memory} (Section~\ref{sec:memory}), which constructs a structured memory pool that stores reusable workflows, subtask skills, and failure patterns as parameterized templates;
\textbf{(2) Memory-Guided Exploration} (Section~\ref{sec:sampling}), which utilizes a stratified group sampling mechanism that injects different strengths of memory guidance into the same GRPO group, facilitating effective advantage estimation while preventing guidance dependency;
\textbf{(3) Self-Evolving Loop} (Section~\ref{sec:evolving}), which continuously refines the memory by extracting novel experience from the newly collected trajectories, enabling progressive improvement and cross-task transfer.

\subsection{Hierarchical Experience Memory}
\label{sec:memory}
Traditional replay buffers store raw trajectories as sequences of state-action pairs, which suffer from high variance and poor generalization to minor UI changes, making it difficult to transfer experience across tasks or apps.
To overcome this limitation, we construct a structured, hierarchical, and abstracted experience memory.
This design allows the agent to retrieve relevant past experience and instantiate it to form specific plans when facing novel tasks.

\paragraph{Memory Representation.}

We draw inspiration from human cognitive processes in GUI navigation, where users naturally decompose tasks into subtasks and rely on familiar interaction patterns from well-known apps. To mimic this, we formulate our memory at three distinct levels:
\begin{itemize}[topsep=0pt, itemsep=3pt, parsep=0pt, partopsep=0pt, after=\vspace{-0.5\baselineskip}]
\item \textbf{High-Level Workflows ($\mathcal{W}$):} At the top level, we store workflow plans that capture global strategies for task completion. Each workflow is an ordered sequence of subtasks. For instance, for the task ``Send an email'', the workflow might be ``[Open Mail App, Select Recipient(s), Type Content, Send]''.
This enables the agent to reuse effective planning for semantically similar tasks.

\item \textbf{Mid-Level Subtask Skills ($\Sigma$):} At the execution level, we maintain a library of subtask skills corresponding to atomic capabilities that recur across many high-level tasks. Examples include ``search for an item'', ``fill a form field'', or ``navigate to a specific folder''. For each subtask, we store summarized action sequences in natural language, such as ``Tap the search icon, type \texttt{\{\{query\}\}}, then select the top result''. These fine-grained skills are highly reusable across applications with similar interaction paradigms.

\item \textbf{Failure Patterns ($\mathcal{F}$):} To prevent repetitive errors, we also record the common failure modes derived from previously failed trajectories. For instance, if an agent frequently fails due to forgetting to enter the correct filename, the memory stores the pattern ``Avoid clicking the `Save' icon before entering a filename''. This enables the agent to avoid known failure modes rather than rediscovering them through costly trial-and-error.

\end{itemize}

\paragraph{Abstraction into Templates.}
A key observation is that many GUI task instructions share a similar structure but differ only in concrete values such as filenames, dates, phone numbers, or UI element names. 
Therefore, we employ an abstraction mechanism that clusters semantically similar instructions and subtask definitions into templates.
Concrete values in the experience are replaced with semantic placeholders. For example, the specific step ``Click the `Save' button to confirm creating report.txt'' becomes the template ``Click the \texttt{\{\{confirm\_button\}\}} button to confirm creating \texttt{\{\{filename\}\}}''. 
This abstraction maximizes transferability by allowing a single parameterized plan to generalize to novel tasks. It also minimizes storage redundancy by consolidating semantically equivalent experiences into compact, easy-to-retrieve templates.

\begin{figure}[t]
  \centering
  \includegraphics[width=1.0\linewidth]{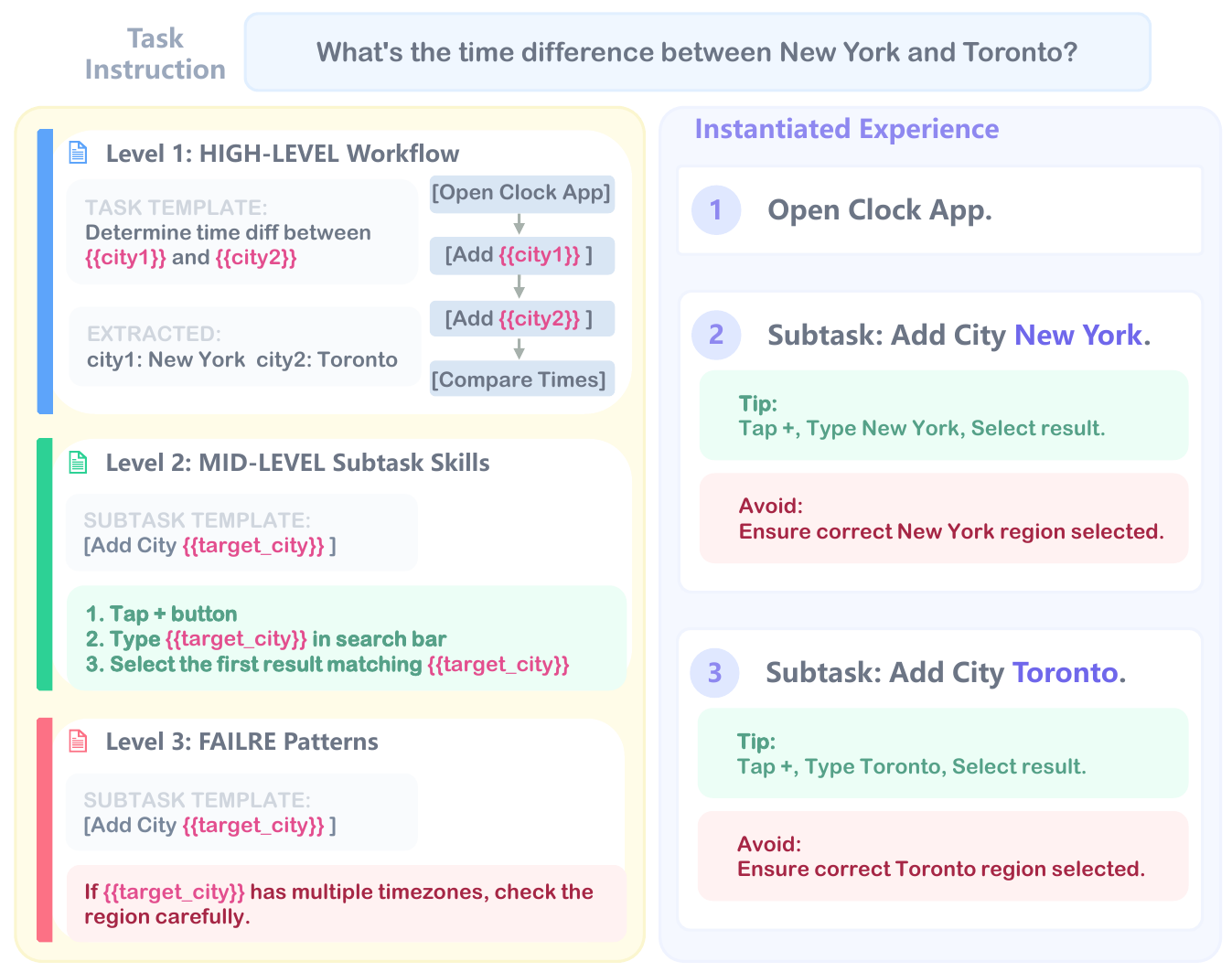}
  \caption{\textbf{Illustration of the Hierarchical Experience Retrieval process.} Given a task instruction, the system performs template matching to extract specific variables (e.g., city names). The retrieved experience is instantiated with the extracted variables to form a concrete, actionable plan for the current rollout.}
  \label{fig:memory_retrieval}
  \vspace{-4mm}
\end{figure}

\paragraph{Storage and Retrieval.}
The memory is stored in a vector database indexed by the embeddings of task and subtask descriptions, enabling efficient similarity search at scale.
The memory retrieval process is illustrated in \cref{fig:memory_retrieval}.
During rollout, given a new task instruction $q$, we first compute its embedding and perform semantic matching with stored task templates. For matched templates, we extract instruction-specific variable bindings and instantiate the corresponding experience by substituting placeholders with the extracted values. 
A retrieved task template may be associated with multiple experience entries accumulated from past explorations.
To balance exploitation and exploration, we employ a scoring mechanism inspired by the Upper Confidence Bound (UCB) algorithm. For a candidate plan $p$, its retrieval score $S(p)$ is:

\begin{equation}
S(p) = \frac{N_{succ}(p)}{\sum_{p' \in \mathcal{P}} N_{succ}(p')} + \lambda_{ucb} \sqrt{\frac{\ln (\sum_{p' \in \mathcal{P}} N_{used}(p'))}{N_{used}(p) + 1}}
\end{equation}
where $N_{succ}$ and $N_{used}$ denote the success count and usage count respectively, and $\lambda_{ucb}$ is the exploration weight. 
The first term encourages the selection of plans with high historical success rates, while the second term provides an exploration bonus for plans that have been tried fewer times. For failure patterns, we use a recency bias to prioritize recent error diagnoses, ensuring that the guidance reflects the agent's latest policy behavior. This adaptive retrieval mechanism ensures that the guidance provided to the agent evolves dynamically as the memory is refined through training.

\subsection{Memory-Guided Exploration}
\label{sec:sampling}

Having established the hierarchical memory structure, we now address how to leverage this memory to enhance online RL training.
A naive approach might simply prepend retrieved experience to every rollout prompt, treating the memory as a fixed reference.
However, this strategy suffers from the risk of dependency overfitting, where the agent learns to follow external instructions rather than internalizing transferable skills.
Furthermore, if strong guidance makes all trajectories in a group successful, the advantage variance becomes zero, hindering optimization in GRPO.
To address these challenges, we propose a stratified group sampling mechanism that injects memory guidance at varying strengths within each rollout batch, combined with a dynamic dropout curriculum that progressively reduces guidance as the agent's ability improves.

\paragraph{Stratified Group Sampling.}
The core insight behind our sampling strategy is that effective GRPO training requires within-group diversity in trajectory outcomes. Rather than applying uniform guidance, we partition each group of $G$ trajectories into three subgroups receiving strong, weak, and no guidance, with respective proportions $\lambda_{\text{strong}}$, $\lambda_{\text{weak}}$, and $\lambda_{\text{none}}$, where $\lambda_{\text{strong}} + \lambda_{\text{weak}} + \lambda_{\text{none}} = 1$. Specifically, for a task $q$, we retrieve the best workflow $\mathcal{W}_q$, skills $\Sigma_q$, and failure patterns $\mathcal{F}_q$. The subgroups are composed as follows:
\begin{itemize}[topsep=0pt, itemsep=3pt, parsep=0pt, partopsep=0pt]
\item \textbf{Strong-Guidance} ($\lambda_{strong} \cdot G$ trajectories): Conditioned on full guidance ($q, \mathcal{W}_q, \Sigma_q, \mathcal{F}_q$). These trajectories exploit the valuable guidance to seek high-quality solutions, thereby providing potential positive anchors to stabilize the learning process.
\item \textbf{Weak-Guidance} ($\lambda_{weak} \cdot G$ trajectories): Conditioned only on the high-level workflow ($q$, $\mathcal{W}_q$). This forces the agent to formulate the low-level execution details, bridging the gap between planning and acting.
\item \textbf{No-Guidance} ($\lambda_{none} \cdot G$ trajectories): Sampled with no external memory, encouraging pure exploration and providing an unbiased estimate of the agent's internalized policy.
\end{itemize}
This stratified design increases the diversity of outcomes within each group, which is essential for estimating advantages in GRPO. Through this mechanism, the unguided policy is trained to match the performance of guided subgroups, thus gradually transferring the knowledge encoded in the external memory into the agent's internal parameters.
\paragraph{Dynamic Dropout Curriculum.}
While stratified group sampling improves within-batch diversity, using fixed guidance proportions throughout training is suboptimal. 
In the early stages of training, strong guidance is crucial for generating positive rewards.
As training progresses, continued reliance can hinder generalization. 
To ensure that the agent eventually internalizes experience, we introduce a dynamic curriculum that gradually withdraws the guidance.
Specifically, we track the Exponential Moving Average (EMA) of the success rate $S_t \in [0, 1]$ for each task. 
We define two thresholds: $\theta_{\text{start}}$, below which the agent is considered to struggle with the task, and $\theta_{\text{end}}$, above which the agent has achieved reliable competence. 
The sampling probabilities $\lambda_{strong}$ and $\lambda_{none}$ are modeled as linear functions of $S_t$:
\begin{equation}
\lambda_{\text{strong}}(S_t) = \text{clip}\left( \lambda_{\text{strong}}^{\max} - \phi(S_t) \Delta \lambda_{\text{strong}}, \lambda_{\text{strong}}^{\min}, \lambda_{\text{strong}}^{\max} \right)
\end{equation}
\begin{equation}
\lambda_{\text{none}}(S_t) = \text{clip}\left( \lambda_{\text{none}}^{\min} + \phi(S_t) \Delta \lambda_{\text{none}}, \lambda_{\text{none}}^{\min}, \lambda_{\text{none}}^{\max} \right)
\end{equation}
where $\phi(S_t) = \frac{S_t - \theta_{\text{start}}}{\theta_{\text{end}} - \theta_{\text{start}}}$ represents the normalized progress, and $\Delta\lambda_{\text{strong}}$, $\Delta\lambda_{\text{none}}$ represent the respective ranges. As the agent's capabilities improve ($S_t \uparrow$), $\lambda_{strong}$ decreases and $\lambda_{none}$ increases, shifting the focus from imitation to independent task solving. 

\paragraph{Guidance-Aware Reward Shaping.}
Standard outcome-based rewards treat all successful trajectories equally, regardless of how the success was achieved. However, in our stratified sampling framework, a success obtained through comprehensive step-by-step guidance is qualitatively less valuable than a success achieved through pure exploration.
To align the reward signal with our optimization objective, we introduce guidance-aware reward shaping that augments the base outcome reward with two components.
First, we calculate a progress reward $r_{progress}$ based on the completion of subtasks defined in the reference workflow $\mathcal{W}$:
\begin{equation}
r_{progress} = \frac{|\mathcal{S}_{completed}|}{|\mathcal{S}_{total}|}
\end{equation}
Second, we add a bonus for unguided success to encourage internalization of skills.
The shaped reward $\mathcal{R}(\tau)$ for a trajectory $\tau$ is computed as:
\begin{equation}
\begin{split}
\mathcal{R}(\tau) = & \lambda_{o} \cdot r_{outcome} + \lambda_{p} \cdot r_{progress} \\
& + \alpha \cdot \mathbb{I}(\text{guidance}(\tau) = \text{None}) \cdot r_{outcome}
\end{split}
\end{equation}
where $\lambda_{o}$ and $\lambda_{p}$ are weights for outcome and progress rewards respectively, $\mathbb{I}(\cdot)$ is the indicator function and $\alpha$ is a bonus coefficient. 
This shaping mechanism provides denser subtask-level learning signals while encouraging the agent to solve tasks without depending on guidance tokens.
\subsection{Self-Evolving Loop}
\label{sec:evolving}
\begin{figure}[t]
  \centering
  \includegraphics[width=1.0\linewidth]{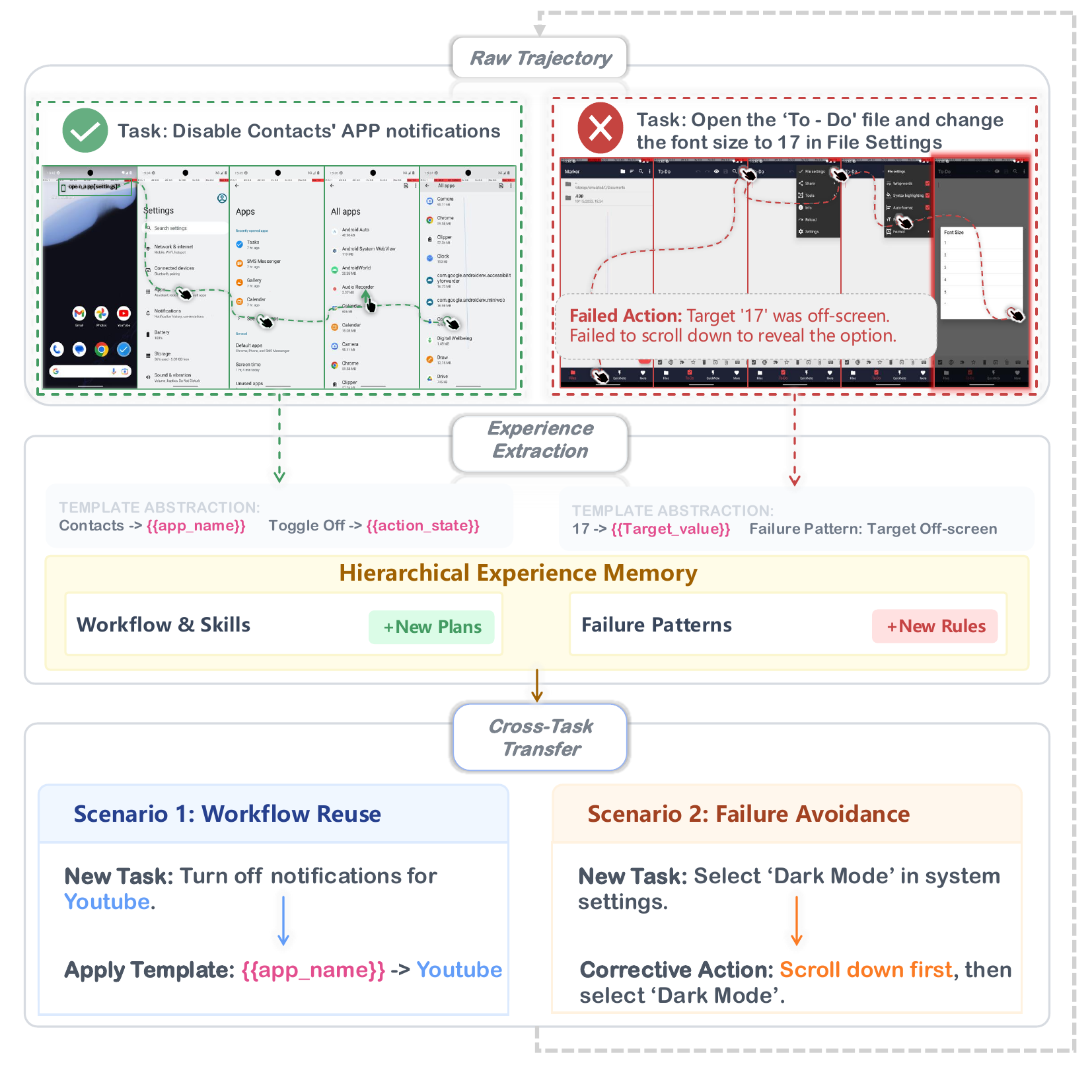}
  \vspace{-4mm}
  \caption{\textbf{The Self-Evolving Loop.} Successful plans and failure causes are extracted from new trajectories to continually refine the memory and guide next rollouts.}
\label{fig:memory_update}
  \vspace{-6mm}
\end{figure}

To ensure that our memory adapts to the evolving policy and captures novel strategies discovered during training, we introduce a self-evolving loop~(\cref{fig:memory_update}) that continuously extracts, abstracts, and integrates new experience from the trajectories generated during online RL.

\paragraph{Experience Extraction and Abstraction.}
At the end of each training iteration, we first employ a reward model to evaluate each trajectory, including determining the global success outcome and identifying the completed subtasks.
Based on these judgments, we invoke an LLM-based experience extraction module to derive structured experience.
For successful trajectories ($\mathcal{R}(\tau)=1$), the module extracts the high-level workflow $\mathcal{W}$ and an execution plan $\Sigma$ for each completed subtask. For failed trajectories ($\mathcal{R}(\tau)=0$), the module extracts valid plans for successfully completed subtasks and generates a failure diagnosis $\mathcal{F}$ specifically for the first failed subtask. To maximize the transferability of extracted experience, we transform these raw extractions into generalized ones by using the abstraction mechanism described in Section~\ref{sec:memory}. 
The extracted workflows, skills, and failure patterns are parameterized by replacing concrete values with semantic placeholders (e.g., replacing ``report.pdf'' with \texttt{\{\{filename\}\}}).

\paragraph{Memory Update.}
Finally, we integrate the abstracted experience into the hierarchical memory $\{\mathcal{W}, \Sigma, \mathcal{F}\}$.
We compute semantic embeddings for the newly extracted entries and query the vector database for existing counterparts with high cosine similarity. 
If a similar entry is found, we merge the new experience by updating the associated statistics: increasing the success count $N_{succ}$ for successful plans or updating the timestamp for failure patterns. If no similar entry exists, we create a new entry with initialized statistics ($N_{succ} = 1$, $N_{used} = 0$).
Through this continual updating process, the memory pool co-evolves with the policy, progressively improving the experience quality and enabling cross-task generalization throughout online training.

\section{Experiments}
\label{sec:experiments}

\subsection{Implementation Details} \label{subsec:exp_setup}

\paragraph{Training Settings.}
To demonstrate the scalability of \ours, we integrate our framework with the Qwen3-VL family~\cite{Qwen3-VL}, specifically the 4B and 8B parameter variants. We implement a parallelized online Reinforcement Learning framework. Specifically, we deploy distributed Android emulator instances across CPU clusters to perform asynchronous rollouts. During the sampling phase, the agent interacts with the environment guided by adaptively retrieved experiences to generate trajectories. We employ the Group Relative Policy Optimization algorithm based on our stratified group sampling. For each query, we sample $G=4$ outputs. We train both models for five epochs with a learning rate of $1 \times 10^{-6}$.

\paragraph{Dataset Construction.}
To construct the training dataset, we collect task instructions from AMEX~\cite{chai2024amex}, AndroidLab~\cite{xu2024androidlab}, and UI-Genie~\cite{xiao2025ui}, filtering for tasks compatible with core apps in our emulator environment. 
To increase task diversity, we employ GPT-4o to augment these seed instructions, resulting in a total of $N_{train}=256$ training queries.
Notably, the original datasets provide only high-level task instructions without explicit subtask definitions. To enable our hierarchical memory structure, we utilize the annotated trajectories to generate subtask definitions, which are subsequently refined through the self-evolving loop during training. 
\paragraph{Reward Evaluation and Experience Extraction.}
A key limitation of prior MLLM-as-a-Judge approaches is that directly processing screenshot sequences often induces severe visual hallucinations. To address this, we adopt a two-stage text-based verification pipeline: we first use Qwen2.5-VL-72B-Instruct~\cite{bai2025qwen2} to convert each screen state and action into textual descriptions, and then employ DeepSeek-V3~\cite{liu2024deepseek} to perform rule-based state verification on the resulting text history. This decoupling of visual grounding from logical reasoning significantly improves evaluation reliability (Accuracy: 0.900 vs. 0.724 for direct MLLM scoring). Then we utilize an experience extraction module (Seed1.8~\cite{seedseed1}) to obtain successful plans and failure diagnoses. Detailed designs and validation results are provided in Appendices~\ref{app:reward_model} and~\ref{app:experience_extraction}.

\subsection{Evaluation Benchmarks} \label{subsec:benchmarks}
We evaluate \ours on two challenging online benchmarks that require dynamic interaction:

\textbf{AndroidWorld}~\cite{rawles2024androidworld} evaluates agents on 116 programmatic tasks across 20 real-world apps. It is characterized by long-horizon dependencies (often $>10$ steps) and dynamic task parameterization, rigorously testing the agent's reasoning and generalization capabilities. Rewards are derived directly from system states, ensuring robust evaluation.

\textbf{AndroidLab}~\cite{xu2024androidlab} focuses on nine commonly used offline applications. It provides fine-grained metrics beyond binary success, making it ideal for analyzing subtask completion and operation efficiency. The benchmark comprises 138 tasks simulating daily user interactions such as managing events, editing notes, and retrieving information.

\subsection{Main Results}

\textbf{Performance on AndroidWorld.}
Table~\ref{tab:androidworld} presents the success rates on AndroidWorld.  \ours demonstrates consistent improvements across model scales.
 The \ours-4B model achieves a success rate of 58.2\%, significantly outperforming the vanilla Qwen3-VL-4B (45.3\%) and even surpassing the larger UI-Venus-7B (49.1\%). This result validates that our memory-guided training enables the agent to learn more efficient policies than standard supervised fine-tuning or vanilla RL. When equipped with the experience memory at inference time (denoted as UI-Mem$^\star$), the performance further improves to 62.5\%. This indicates that \ours also learns to effectively retrieve and utilize external memory to solve novel tasks.
 On the 8B scale, \ours-8B with memory retrieval achieves a state-of-the-art success rate of 71.1\%, outperforming closed-source commercial APIs such as Gemini-2.5-Pro and Seed1.8. This suggests that our hierarchical memory mechanism scales effectively with model capacity, unlocking stronger reasoning capabilities in larger foundation models.

 \begin{table}
  \caption{\textbf{Performance comparison on the AndroidWorld benchmark.} The table presents closed-source API models, open-source foundation models and our approach. $^\star$ denotes inference-time memory retrieval. Best results are in \textbf{bold}, and the second-best results are \underline{underlined}.}
  \label{tab:androidworld}
  \centering
  \small
  \resizebox{1.0\linewidth}{!}{
  \begin{tabular}{l c c}
    \toprule
    \textbf{Model} & \textbf{Params.} & \textbf{Success Rate} \\
    \midrule
    Seed1.5-VL~\cite{guo2025seed1} & - & 62.1 \\
     UI-Tars-1.5~\cite{ui-tars-15-seed} & - & 64.2 \\
    Gemini-2.5-Pro~\cite{comanici2025gemini} & - & \underline{69.7} \\
    Seed1.8~\cite{seedseed1} & - & \textbf{70.7} \\
    \midrule
     Qwen3-VL-2B~\cite{Qwen3-VL}  & 2B & 36.4 \\
    MAI-UI-2B~\cite{zhou2025mai} & 2B & 49.1 \\
    ScaleCUA-3B~\cite{liu2025scalecua}  & 3B & 23.7 \\
     Ferret-UI Lite-3B~\cite{yang2025ferret}  & 3B & 28.0 \\
    
    Qwen3-VL-4B~\cite{Qwen3-VL}  & 4B & 45.3  \\
   \rowcolor{gray!15} \ours-4B  & 4B & \underline{58.2} \\
    \rowcolor{gray!15} \ours-4B$^\star$  & 4B & \textbf{62.5} \\
    \midrule
    UI-Venus-7B~\cite{gu2025ui}  & 7B & 49.1 \\
     GUI-Owl-7B~\cite{ye2025mobile} & 7B & 66.4 \\
    Step-GUI-8B~\cite{yan2025step}  & 8B & \underline{67.7} \\
     UI-Tars-1.5-7B~\cite{ui-tars-15-seed}  & 7B & 30.0 \\
     Qwen3-VL-8B~\cite{Qwen3-VL}  & 8B & 47.6 \\
    \rowcolor{gray!15} \ours-8B  & 8B &  66.8 \\
    \rowcolor{gray!15} \ours-8B$^\star$  & 8B &  \textbf{71.1} \\
    
    \bottomrule
  \end{tabular}
  }
  \vspace{-6mm}
\end{table}

\textbf{Performance on AndroidLab.}
We utilize AndroidLab to analyze fine-grained capabilities using Sub-Goal Success Rate (Sub-SR), Reversed Redundancy Ratio (RRR), Reasonable Operation Ratio (ROR), and overall Success Rate (SR). Results are shown in Table~\ref{tab:androidlab}. \ours shows a substantial improvement in Sub-SR compared to the vanilla baseline, demonstrating the effectiveness of our hierarchical memory. By explicitly retrieving workflows and subtask plans, the agent can decompose complex long-horizon tasks into manageable segments, thereby reducing error propagation. Notably, \ours-8B outperforms MobileRL (42.5\%), a strong baseline that utilizes static experience replay. While MobileRL relies on the reuse of successful trajectories, \ours employs active memory to guide the generation of high-quality new trajectories with diverse progress rewards. This allows our agent to explore the state space more efficiently and adapt to dynamic UI changes, rather than merely memorizing past paths.

\begin{table*}[h]
  \caption{\textbf{Performance comparison on the AndroidLab benchmark.} The table presents closed-source API models, open-source foundation models and our approach. $^\star$ denotes inference-time memory retrieval. The best results are in \textbf{bold}, and the second-best are \underline{underlined}.}
  \label{tab:androidlab}
  \centering
  \small
  \resizebox{0.95\linewidth}{!}{
  \begin{tabular}{l C{1.5cm} C{2.5cm} C{2.5cm} C{2.5cm} C{2cm}}
    \toprule
    Model & Params & Sub-Goal\newline Success Rate & Reversed Redun-\newline dancy Ratio & Reasonable\newline Operation Ratio & Success Rate \\
    \midrule
    Gemini-1.0              & -- & 12.6 & 72.5 & 76.7 & 10.9 \\
    Claude-3-Opus           & -- & 15.1 & 81.4 & 83.9 & 13.0 \\
    Gemini-1.5-Pro          & -- & 18.5 & \underline{106.0} & \textbf{91.5} & 16.7 \\
    Claude-3.5-Sonnet       & -- & \underline{32.7} & \textbf{113.4} & 81.2 & 29.0 \\
    GPT-4o                  & -- & \textbf{35.0} & 87.3 & \underline{85.4} & \underline{31.2} \\
    AutoGLM                 & -- & --   & --    & --    & \textbf{36.2} \\
    \midrule
    UI-Genie-Agent  & 3B & 35.4 & 91.3 & 90.6 & 28.8 \\
    Qwen3-VL-4B  & 4B & 48.2 & \textbf{93.0} & 90.5 & 37.0 \\
    \rowcolor{gray!15} \ours-4B  & 4B & \underline{49.5} &	\textbf{93.0} &	\underline{93.5} & \underline{37.7} \\
    \rowcolor{gray!15} \ours-4B$^\star$  & 4B & \textbf{51.9} & 89.2 & \textbf{94.6} &	\textbf{39.9} \\

    \midrule
    
    LLaMA3.2-11B-Vision     & 11B & 13.0 & 61.7 & 87.9 & 10.1 \\
    CogVLM2-ft              & 19B & 16.1 & 57.4 & 85.6 & 11.6 \\
    Qwen2.5-VL          & 7B & 18.7 & 70.6 & 76.8 & 14.9 \\
    UI-Genie-Agent  & 7B & 46.3 & 92.8 & 91.4 & 38.7 \\
    Qwen3-VL-8B  & 8B & 45.3 & \textbf{97.7} & 91.8 & 34.8 \\
    UI-TARS-1.5-7B & 7B & 49.4 & 84.2 & \underline{92.5} & 40.6\\
    MobileRL & 7B & - & - & - & 42.5\\
    
   \rowcolor{gray!15} \ours-8B  & 8B &  \underline{52.7} & \underline{93.9} & 90.9 & \underline{43.5} \\
    \rowcolor{gray!15} \ours-8B$^\star$  & 8B & \textbf{56.0} & 89.8 & \textbf{94.9} & \textbf{44.9} \\

    \bottomrule
  \end{tabular}
  }
   \vspace{-2mm}
\end{table*}

\subsection{Ablation Study}

\paragraph{Comparison with RL Training Paradigms.}
We compare our method with other RL methods to demonstrate the advantage of our memory-guided approach. As shown in Table~\ref{tab:ablation_rl}, we compare \ours with: (1) Vanilla GRPO, which adopts GRPO training from scratch without external memory; (2) GRPO + Progress Reward, which augments GRPO with subtask-level dense rewards; (3) Inference-time Prompting, which retrieves experience from our memory as context during inference only; (4) GRPO + Experience Replay, which mixes top-k past successful trajectories into the training batch; and (5) GRPO + Inference-time Prompting, where we apply memory retrieval to the GRPO-trained model. Results indicate that vanilla GRPO struggles with exploration in the vast GUI state space. While dense rewards improve the performance (57.3\%), they cannot transfer past experience to novel tasks. Experience Replay improves training stability (56.5\%), but remains limited to replaying static historical data and fails to actively guide exploration toward new solutions. Notably, Inference-time Prompting yields limited gains when applied to the baseline and GRPO-trained model, suggesting that simply providing context is insufficient if the model has not internalized how to utilize it. In contrast, \ours achieves the highest success rate, demonstrating that our approach, which combines memory-guided exploration during training with adaptive retrieval, substantially outperforms other methods.
 
\begin{table}[h]
\caption{\textbf{Comparison of different RL training paradigms on AndroidWorld.}}
\label{tab:ablation_rl}
\centering
  \resizebox{1.0\linewidth}{!}{
\begin{tabular}{l c c}
\toprule
\textbf{Method} & \textbf{Mechanism} & \textbf{SR (\%)} \\
\midrule
Qwen3-VL-8B  & Baseline & 47.6 \\
\midrule
Vanilla GRPO & Online RL & 53.0 \\
GRPO + Progress Reward & Online RL & 57.3 \\
Inference-time Prompting & RAG (No Training) & 52.6 \\
GRPO + Experience Replay & Data Reuse & 56.5 \\
GRPO + Inference-time Prompting & Online RL+RAG & 55.2 \\
\midrule
\textbf{\ours (Ours)} & \textbf{Memory-Guided RL} & \textbf{66.8} \\
\bottomrule
\end{tabular}
}
\vspace{-2mm}
\end{table}

\paragraph{Component Analysis.}

\begin{figure}[t]
    \centering
    \includegraphics[width=0.98\linewidth]{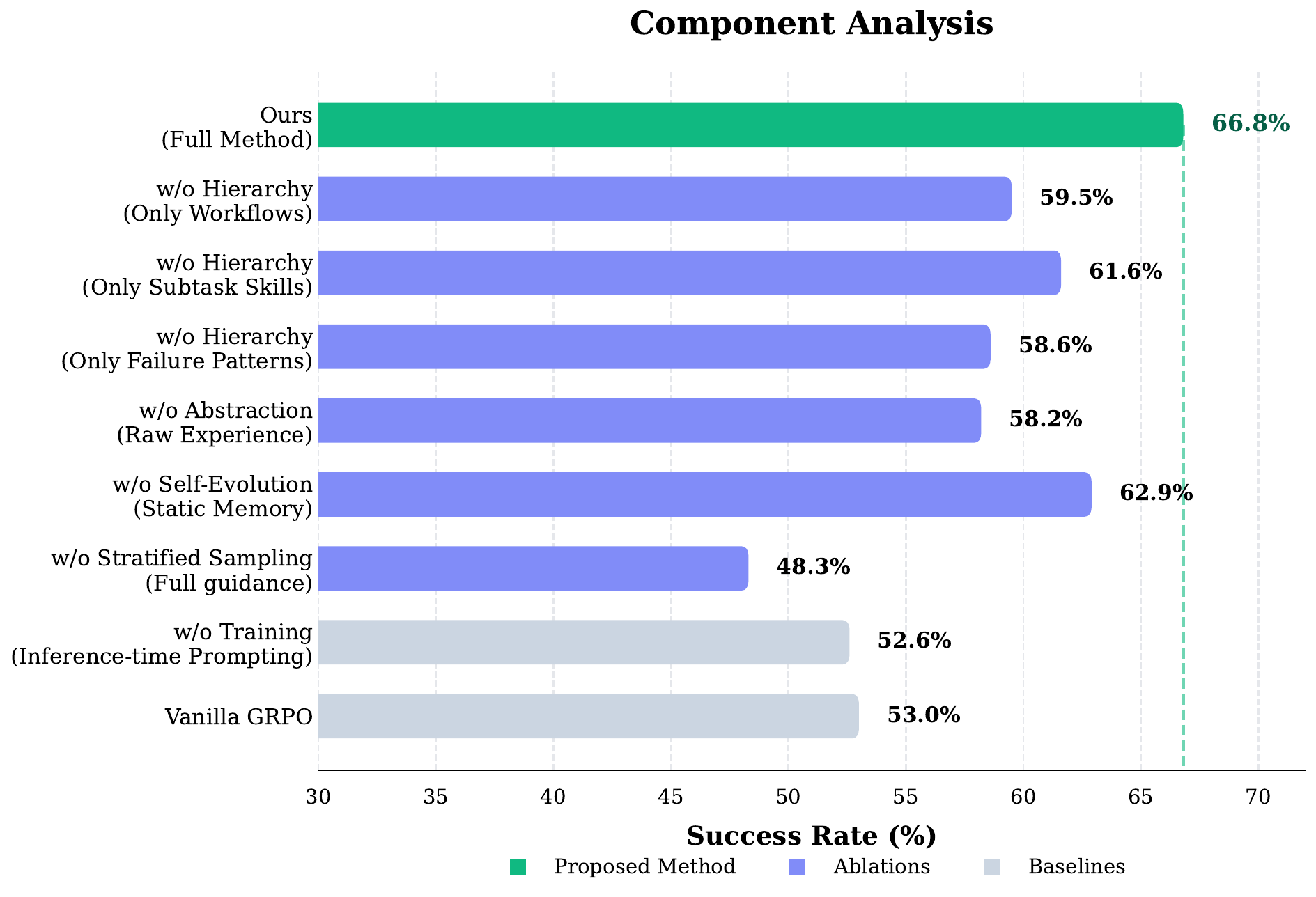}
    \vspace{-1mm}
    \caption{\textbf{Component analysis of \ours.} We evaluate the impact of removing different components in our framework.}
    \label{fig:ablation_chart}
    \vspace{-2mm}
\end{figure}

We further evaluate the contribution of different components in our framework. As illustrated in Figure~\ref{fig:ablation_chart}, we compare the full \ours framework against variants where specific components are removed or revised.
We first analyze the effectiveness of our hierarchical memory structure. Removing the hierarchy to rely solely on specific components leads to significant performance drops. Using only high-level workflows, low-level subtask skills or failure patterns each underperforms the full method by a significant margin. This demonstrates that the three experience types of our memory system capture complementary knowledge.
Moreover, replacing our abstracted memory with raw experience results in a significant performance drop (58.2\%).
This confirms that raw data are verbose, task-specific, and difficult to generalize across different tasks.
Furthermore, disabling the memory update reduces performance to 62.9\%, highlighting the importance of continuously refining the memory pool during the training progress.
Providing full memory guidance without stratified sampling significantly decreases the performance. 
This result reveals that excessive guidance can lead to over-reliance on memory, undermining the agent's ability to develop autonomous reasoning.
Our stratified sampling strategy effectively balances guidance and exploration, enabling the agent to benefit from memory without sacrificing its capacity for independent problem-solving.

\begin{figure}[t]
    \centering
    \includegraphics[width=0.98\linewidth]{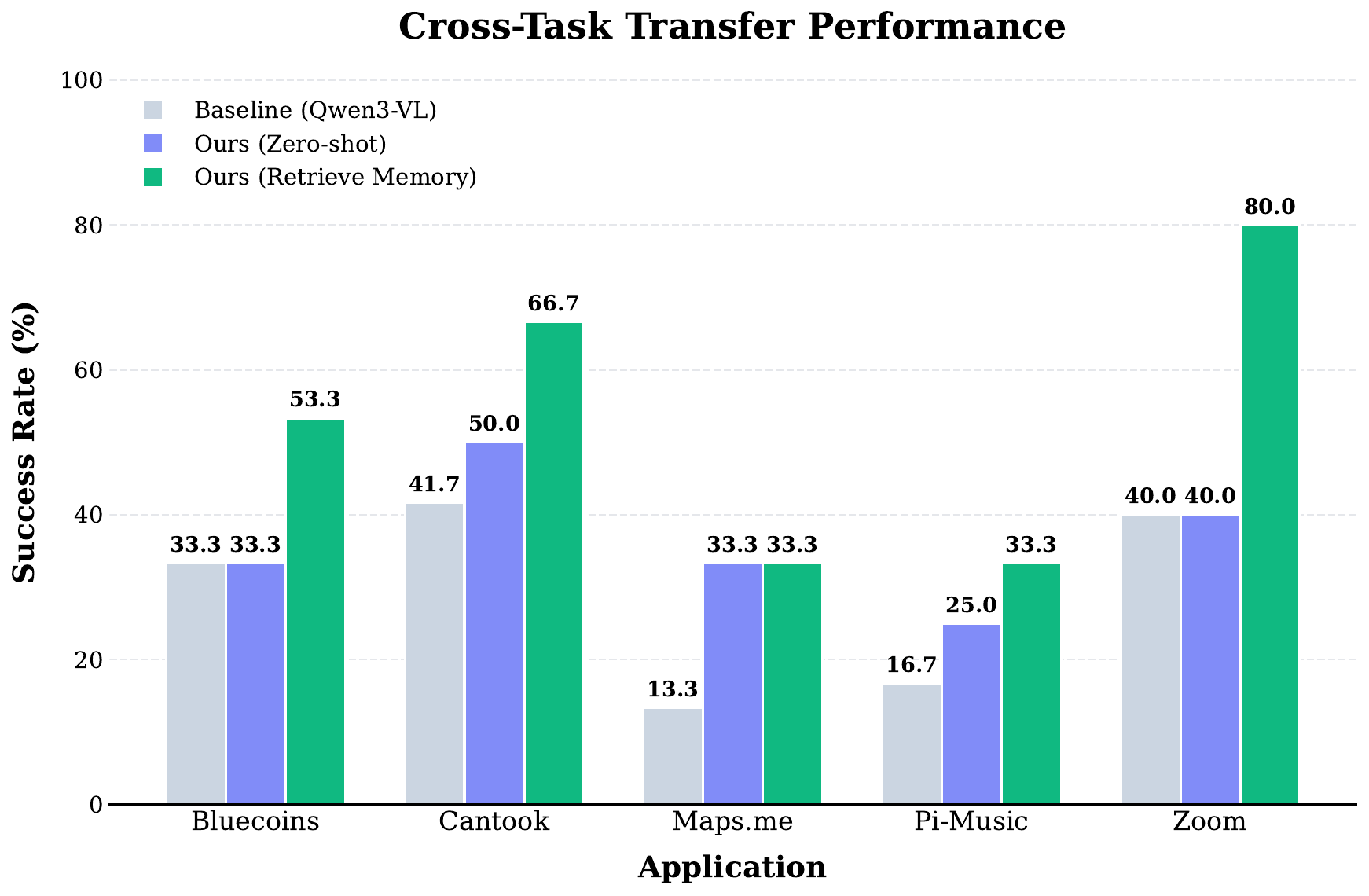}
    \vspace{-1mm}
    \caption{\textbf{Cross-task generalization performance on five held-out applications unseen during training.}}
    \label{fig:cross_task}
\end{figure}

\paragraph{Cross-Application Generalization.}
While our main benchmarks (AndroidWorld and AndroidLab) already evaluate out-of-domain tasks, we conduct a stricter generalization test by evaluating on applications entirely unseen during training. Specifically, we select five held-out apps from AndroidLab that are completely absent from our training set: Bluecoins, Cantook, Maps.me, Pi-Music, and Zoom. 
Figure~\ref{fig:cross_task} compares the success rate of the baseline model, our method applied in a zero-shot manner (without memory retrieval), and our full method with retrieved memory. As shown, our zero-shot model already outperforms or matches the baseline across all five apps, demonstrating that our training method improves generalization even without explicit memory support. More notably, incorporating retrieved memory at inference time provides consistent additional gains, with particularly substantial improvements on Bluecoins and Maps.me. This indicates that experiences learned from training apps can effectively transfer to novel applications through our memory retrieval mechanism. 

\begin{figure}[t]
    \centering
    \includegraphics[width=0.98\linewidth]{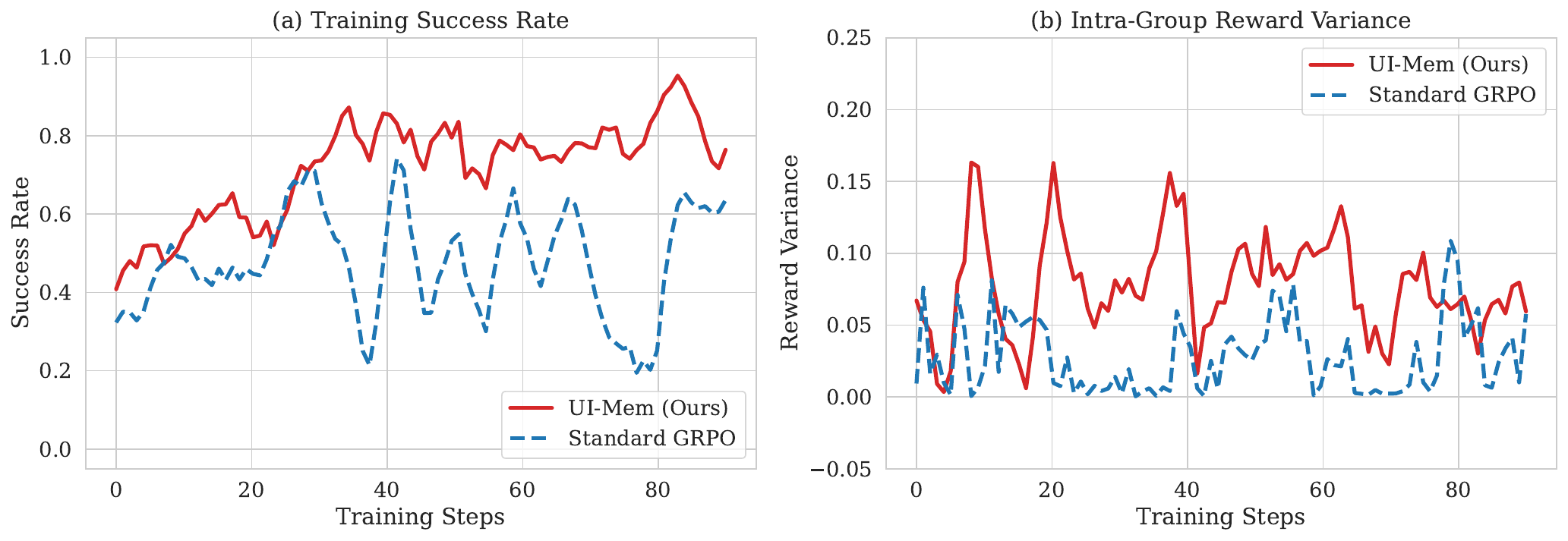}
    \caption{\textbf{Training dynamics comparison between \ours and standard GRPO.} (a) Success rate over training steps. (b) Mean intra-group reward variance, which directly impacts the effectiveness of advantage estimation.}
    \label{fig:ablation_curves}
    \vspace{-6mm}
\end{figure}

\paragraph{Impact of Memory-Guided Exploration.}
To validate the effectiveness of our framework, we analyze the training dynamics in Figure~\ref{fig:ablation_curves}. As shown in Figure~\ref{fig:ablation_curves}(a), standard GRPO exhibits highly unstable training dynamics, with success rate fluctuating dramatically throughout the training process without showing a clear upward trend. In contrast, \ours achieves stable improvement and faster convergence. This performance gap is mechanistically explained by Figure~\ref{fig:ablation_curves}(b), which plots the intra-group reward variance—a critical factor for advantage estimation in GRPO. 
The baseline frequently collapses to near-zero when sampled groups consist entirely of failures, resulting in vanishing gradient.
The baseline exhibits negligible variance (near zero) as most sampled groups consist entirely of failures, resulting in vanishing gradients. Conversely, our Stratified Group Sampling ensures a mixture of successful (guided) and exploratory trajectories within each group, maintaining a healthy variance level. This continuous supply of informative gradients facilitates effective policy optimization and progressive internalization of external memory into the agent's parameters.

\section{Conclusion}
In this work, we introduced \ours, a novel framework designed to overcome the fundamental inefficiencies of online Reinforcement Learning in GUI environments. \ours constructs a hierarchical, self-evolving memory that decomposes raw experiences into reusable workflows, subtask skills, and failure patterns. We utilized this memory through a stratified group sampling mechanism tailored for GRPO, which balances memory-guided exploitation with necessary exploration to facilitate effective advantage estimation. We also introduced a self-evolving loop to continually refine the memory pool to adapt to the current policy.
Experiments on challenging GUI benchmarks demonstrate that \ours significantly outperforms baselines in both sample efficiency and success rate, with strong cross-task generalization enabled by reusable experience transfer.
\section*{Impact Statement}

This paper presents work whose goal is to advance the field of Machine Learning, specifically focusing on the efficiency and generalization of autonomous GUI agents. This work can assist users with disabilities or limited technical proficiency in navigating complex mobile interfaces with natural language commands. However, we acknowledge several potential risks associated with this research. The deployment of Online RL agents involves direct interaction with live environments. During the exploration phase, an agent might perform unintended actions (e.g., deleting data or financial transactions), necessitating the development of robust safe exploration protocols before real-world deployment. 
Moreover, while our memory abstraction mechanism replaces sensitive text with placeholders like \texttt{\{\{password\}\}}, processing screenshots with personal data still raises privacy concerns that demand strict handling procedures.


\bibliography{example_paper}
\bibliographystyle{icml2026}

\newpage
\appendix
\onecolumn
\section*{Appendix}
\section{Preliminaries}
\label{app:pre}
\paragraph{GUI Task Formulation.} 
We formalize the interaction between the MLLM agent and the Graphical User Interface (GUI) as a finite-horizon Markov Decision Process (MDP), represented by the tuple $\mathcal{M} = \langle \mathcal{S}, \mathcal{A}, \mathcal{P}, \mathcal{R}, H \rangle$. 
The state space $\mathcal{S}$ captures the multi-modal context; a state $s_t \in \mathcal{S}$ comprises the pixel-level screenshot $I_t$ and the high-level user instruction $q$. 
The action space $\mathcal{A}$ consists of atomic GUI operations (e.g., \texttt{Click}, \texttt{Swipe}, \texttt{Type}, \texttt{Home}).
At timestep $t$, the agent policy $\pi_\theta(a_t | s_t, q)$ generates an action $a_t$, transitioning the environment to $s_{t+1}$ according to the system dynamics $\mathcal{P}(s_{t+1} | s_t, a_t)$. 
Reflecting the sparse nature of real-world GUI tasks, the reward function is defined as $\mathcal{R}(\tau) = r_{outcome}\in \{0, 1\}$, provided only at the termination of the trajectory $\tau = (s_0, a_0, \dots, s_T, a_T)$, where $T \le H$. 

\paragraph{Group Relative Policy Optimization (GRPO).}
We adopt GRPO as our base optimization algorithm, as it eliminates the need for a separate critic model. GRPO samples a group of $G$ trajectories $\{\tau_i\}_{i=1}^G$ for a task instruction $q$ from the current policy $\pi_{\theta_{old}}$. The advantage $A_i$ for each trajectory is computed by normalizing rewards within the group:
\begin{equation}
    A_i = \frac{\mathcal{R}(\tau_i) - \text{mean}(\{\mathcal{R}(\tau_j)\}_{j=1}^G)}{\text{std}(\{\mathcal{R}(\tau_j)\}_{j=1}^G) + \epsilon}
\end{equation}
where $\epsilon$ is a small constant for numerical stability. The policy is updated by maximizing the following objective:

\begin{equation}
\begin{split}
    & \mathcal{J}_{GRPO}(\theta) = \mathbb{E}_{c \sim \mathcal{D}, \{\tau_i\} \sim \pi_{\theta_{old}}} \Bigg[  \frac{1}{G} \sum_{i=1}^G \min \bigg( \frac{\pi_\theta(\tau_i)}{\pi_{\theta_{old}}(\tau_i)} A_i, \\
    & \text{clip}\left(\frac{\pi_\theta(\tau_i)}{\pi_{\theta_{old}}(\tau_i)}, 1-\delta, 1+\delta\right) A_i \bigg) 
     - \beta \mathbb{D}_{KL}(\pi_\theta || \pi_{ref}) \Bigg]
\end{split}
\end{equation}

where $\delta$ is the clipping hyperparameter and $\beta$ controls the KL-divergence penalty from the reference model $\pi_{ref}$ to prevent policy collapse.

\section{Example of Hierarchical Memory Structure}
We provide a detailed example of our hierarchical memory representation structure in Figure~\ref{fig:data_example}. Beyond standard task instructions, we introduce an \texttt{essential\_states\_template}, which defines the key intermediate states that characterize the valid completion of the overall task, thereby facilitating our reward evaluation based on rule-based state verification.

\begin{figure*}[t]
    \centering
    \begin{minipage}{0.9\linewidth}
\begin{lstlisting}[language=json]
[
  {
    "task_id": "AutoGenerated_Task_75",
    "package_name": ["net.gsantner.markor"],
    "task_template": {
      "content": "Rename file {{current_filename}} to {{new_filename}} in Markor.",
      "parameter_config": {
        "fixed_parameters": {
          "app_package": "net.gsantner.markor",
          "rename_icon_name": "A",
          "confirm_button_name": "OK"
        },
        "variable_parameters": ["current_filename", "new_filename"]
      }
    },
    "essential_states_template": {
      "S1": { 
        "content": "File '{{current_filename}}' is selected with context options visible.",
        "variable_mapping": { "current_filename": "current_filename" }
      },
      "S2": { "content": "Rename dialog is active." },
      "S3": { 
        "content": "File is renamed to '{{new_filename}}'.",
        "variable_mapping": { "new_filename": "new_filename" }
      }
    },
    "subtask_template": {
      "T1": {
        "subtask_label": "Select Note via Long Press",
        "content": "Long press the file named '{{current_filename}}'."
      },
      "T2": {
        "subtask_label": "Tap Rename Icon",
        "content": "Tap the '{{icon_name}}' icon to open rename options."
      },
      "T3": {
        "subtask_label": "Enter Text and Confirm",
        "content": "Enter '{{new_filename}}' and tap '{{button_name}}' to confirm."
      }
    }
  }
]
\end{lstlisting}
    \end{minipage}
    \caption{\textbf{Example of Hierarchical Data Structure.} The \texttt{essential\_states\_template} defines the critical milestones of the task, providing a mechanism for state-based verification rather than directly judging subtask or task completion.}
    \label{fig:data_example}
\end{figure*}

\section{Reward Model Details}
\label{app:reward_model}
\paragraph{Reward Model Design.}
We identify a significant limitation in previous MLLM-as-a-Judge approaches: directly feeding the entire history of screenshots often leads to severe hallucination issues. To address this, we propose a text-based verification approach. 

\begin{figure*}[t]
    \centering
    \begin{minipage}{0.48\linewidth}
        \centering
        \includegraphics[width=\linewidth]{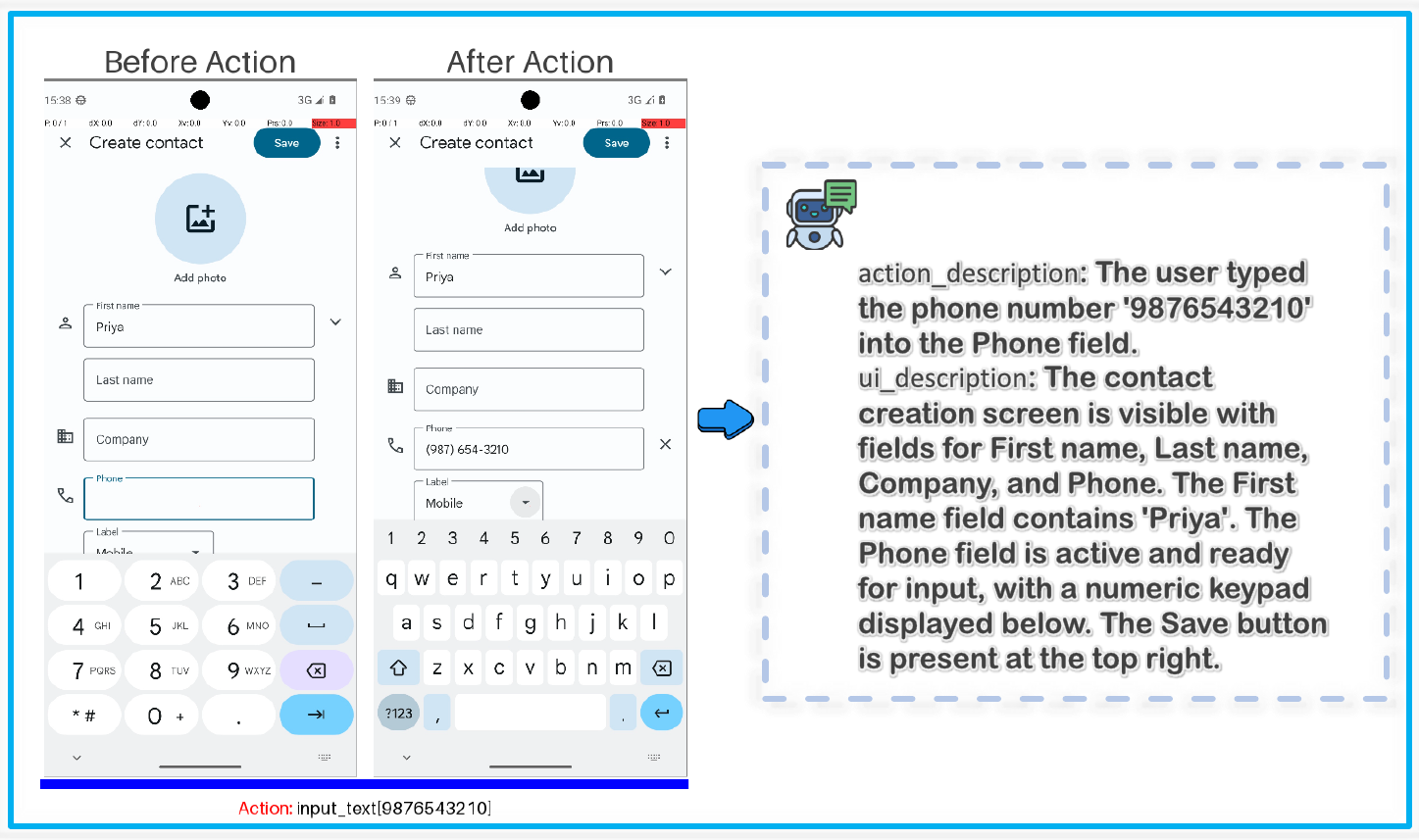}
        \centerline{\footnotesize (a) Text Input Operation}
        \label{fig:action_desc_1}
    \end{minipage}
    \hfill
    \begin{minipage}{0.48\linewidth}
        \centering
        \includegraphics[width=\linewidth]{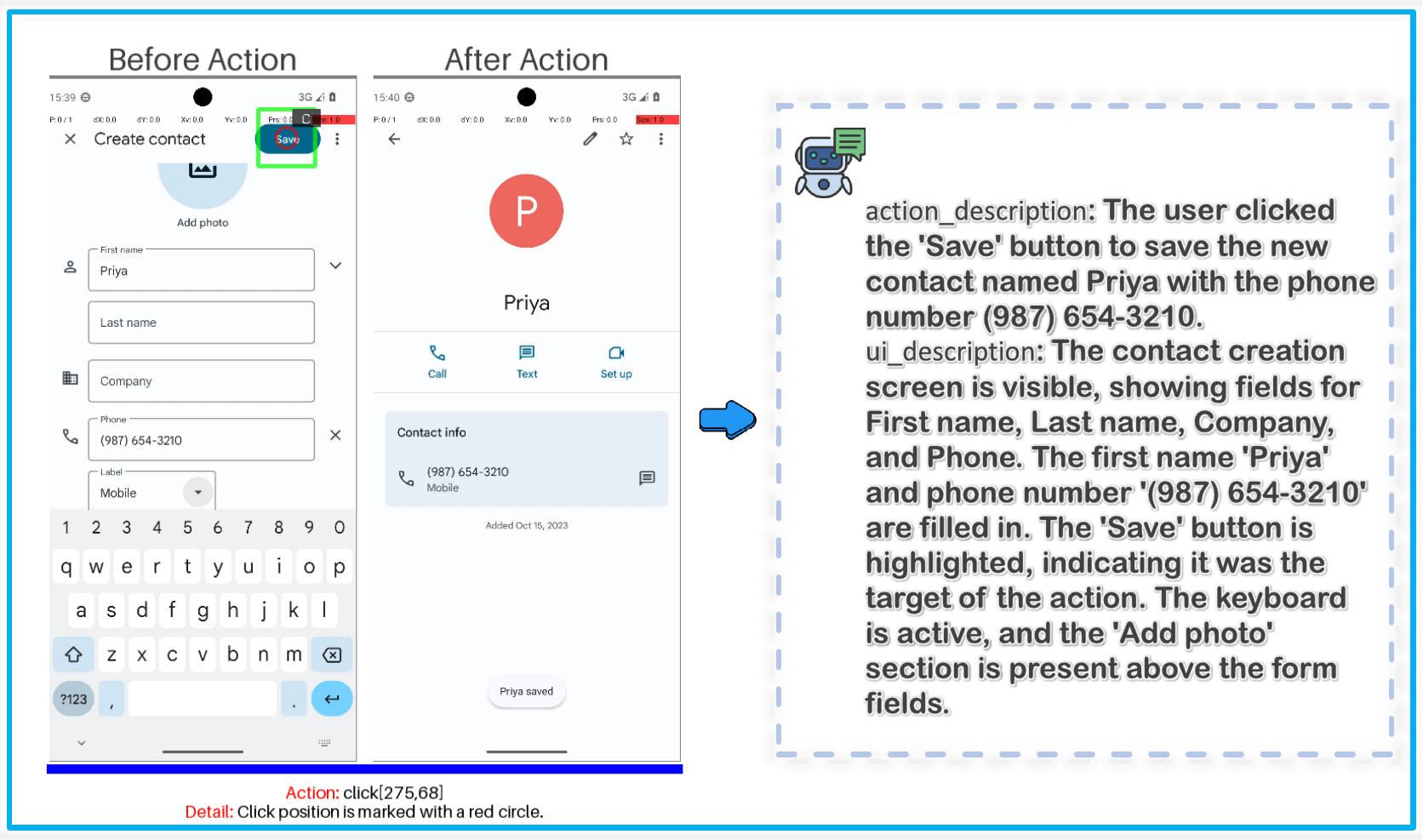}
        \centerline{\footnotesize (b) Button Click Operation}
        \label{fig:action_desc_2}
    \end{minipage}
    \caption{Textual Description Examples of UI states and executed actions. By comparing \textit{Before Action} and \textit{After Action} screens, the MLLM provides (1) \texttt{ui\_description}, capturing the visual state and element visibility relevant to the current task, and (2) \texttt{action\_description}, specifying the actual operation executed (e.g., typing text, clicking buttons). (a) shows a \textbf{Text Input} scenario where the user types a phone number, capturing the input field state change. (b) illustrates a \textbf{Button Click} operation (saving a contact), identifying the interactable element. These structured descriptions serve as grounded inputs for the Reward Computation.}
    \label{fig:action_desc_gen}
\end{figure*}

\begin{figure*}[t]
    \centering
    \includegraphics[width=\linewidth]{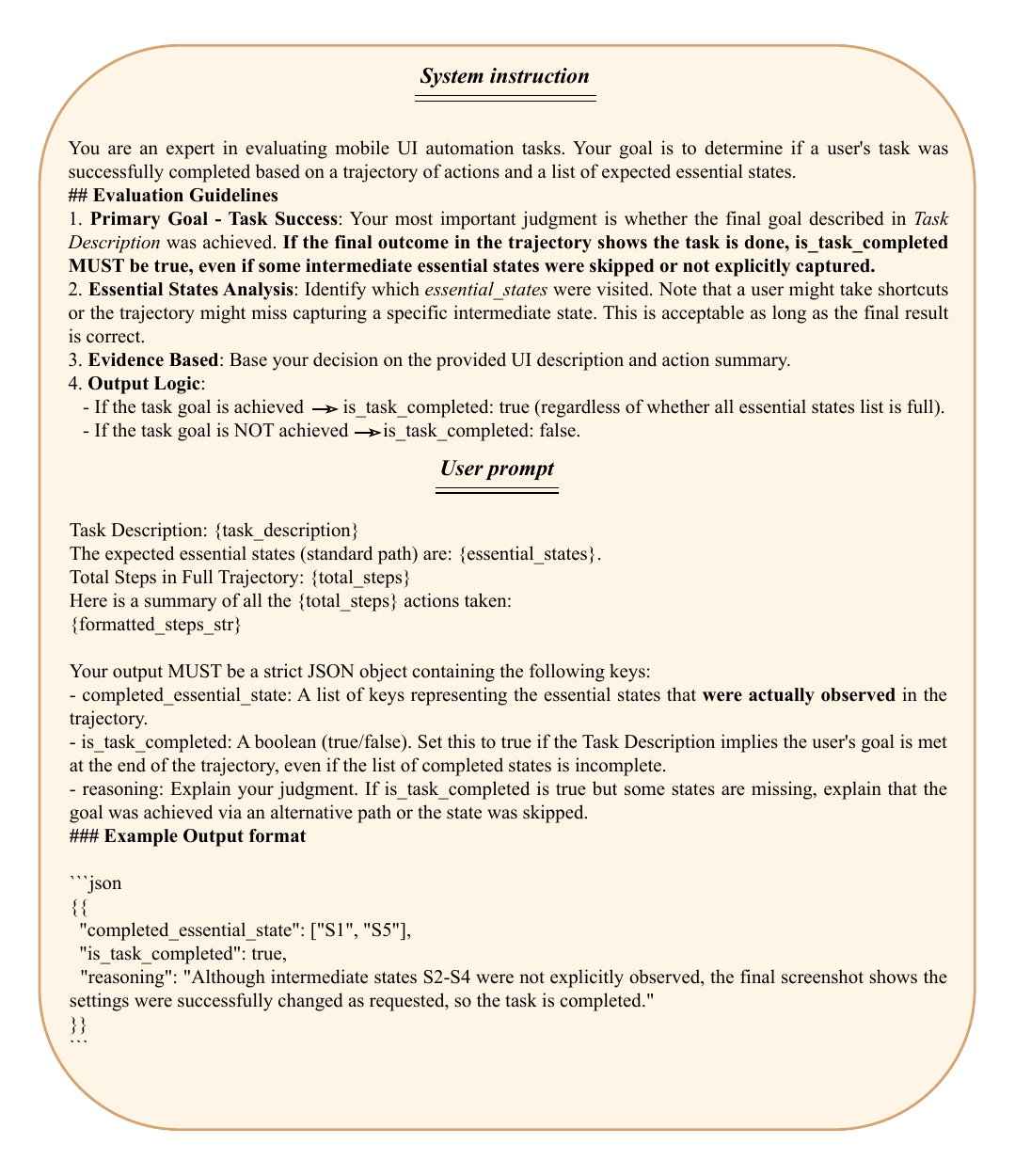}
    \caption{Prompt design for the LLM verification Model.}
    \label{fig:reward_prompt}
\end{figure*}

Our method operates in two stages. First, we utilize Qwen2.5-VL-72B-Instruct to generate textual descriptions for each screen state and the corresponding executed actions in the trajectory. As illustrated in Figure~\ref{fig:action_desc_gen}, by analyzing paired screenshots captured before and after an action, the model provides precise descriptions of the UI elements relevant to the task instruction, as well as the specific operation performed. These descriptions are then concatenated to form a textual history of the agent's actual actions. After that, we feed this textual history into an open-source LLM, DeepSeek-V3, which is prompted to identify which essential states are completed based on the UI descriptions and action descriptions. This effectively converts the previous visual-based evaluation into a robust rule-based verification process, which we have found to be significantly more accurate than direct MLLM evaluation. Based on the model response, we calculate a precise progress reward $\mathcal{R}_{progress} = |\mathcal{S}_{completed}|/|\mathcal{S}_{total}|$, offering a dense signal for optimization. By grounding reasoning in verified state changes rather than raw pixels, this approach significantly reduces the rate of hallucination common in pure MLLM-based evaluators.
We provide our prompt for the LLM verification model in Figure~\ref{fig:reward_prompt}. 

\paragraph{Reward Model Evaluation.}
We quantitatively assess the accuracy of our reward computation method and compare it against previous MLLM-as-a-Judge methods. Table~\ref{tab:reward_model_valid} reports the prediction accuracy, Precision, Recall, and F1 Score across different backbone models. The baseline MLLM-based approaches often struggle with visual hallucinations, resulting in suboptimal F1 scores (e.g., 0.837 for Gemini 2.5 Pro). In contrast, our proposed method in \ours framework, which employs a two-stage text-based verification pipeline, consistently achieves superior performance. Notably, the Qwen2.5-VL-72B-Instruct+DeepSeek-V3 settings reaches an F1 score of 0.902, significantly outperforming the MLLM-as-a-Judge method of using closed-source API model Gemini 2.5 Pro. This demonstrates that decoupling visual grounding from logical verification effectively mitigates hallucinations, providing a robust and reliable reward signal for reinforcement learning. We adopt this settings for our reward evaluation during online training.

\begin{table*}[t]
  \caption{Reward Model Accuracy Comparison. \ours Text-based Verification based on open-source models (Qwen2.5-VL-72B-Instruct + DeepSeek-V3) achieves competitive performance compared with strong closed-source API models across all metrics.}
  \label{tab:reward_model_valid}
  \centering
  \small
  \resizebox{1.0\linewidth}{!}{
  \begin{tabular}{l l c c c c}
    \toprule
    & \textbf{Model} & \textbf{Accuracy} & \textbf{Precision} & \textbf{Recall} & \textbf{F1 Score} \\
    \midrule
    \multirow{3}{*}{MLLM-as-a-Judge} & Gemini 2.5 Pro & 0.776 & 0.845 & 0.908 & 0.837 \\
    & Seed1.5-VL  & 0.639 & 0.879 & 0.770 & 0.775 \\
    & Qwen2.5-VL-72B-Instruct & 0.724 & 0.841 & 0.845 & 0.811 \\
    \midrule
    \multirow{3}{*}{\ours (Text-based)} & Gemini 2.5 Pro + DeepSeek-V3 & \textbf{0.922} & 0.912 & 0.967 & \textbf{0.916} \\
    & Seed1.5-VL + DeepSeek-V3 & 0.880 & 0.907 & 0.921 & 0.887 \\
    & Qwen2.5-VL-72B-Instruct + DeepSeek-V3 & \underline{0.900} & 0.909 & 0.943 & \underline{0.902} \\
    \bottomrule
  \end{tabular}
  }
\end{table*}

\section{Experience Extraction Details}
\label{app:experience_extraction}
We employ an experience extraction model, specifically Seed1.8, due to its strong reasoning abilities, to derive both success plans and failure patterns from newly generated trajectories. Based on the evaluation results provided by our reward model, this extraction process is divided into two distinct phases: Subtask-Level Experience Extraction and High-level Workflow Extraction.

\begin{figure*}[t]
    \centering
    \includegraphics[width=\linewidth]{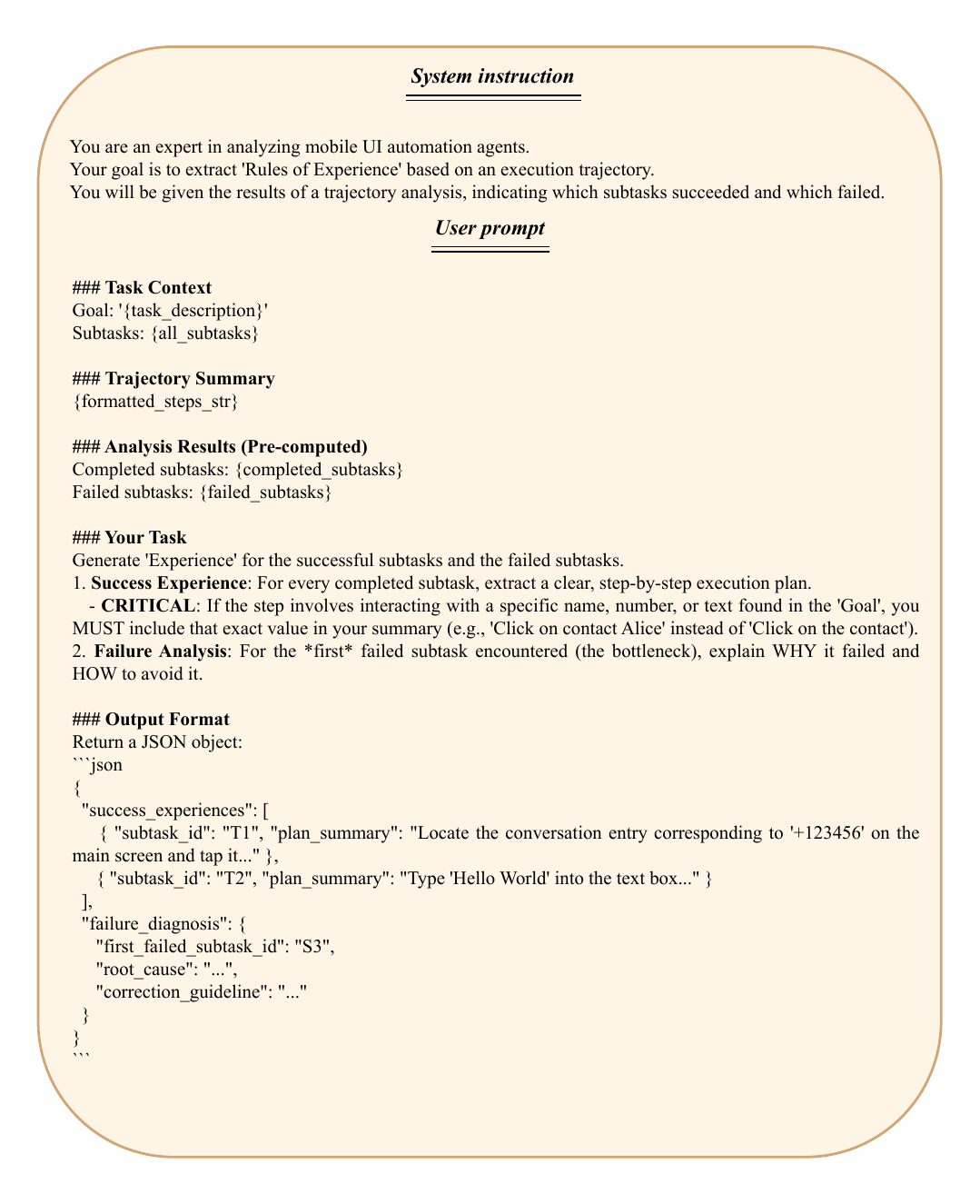}
    \caption{Prompt design for Experience Extraction for subtask-level skills and failure analysis. This module identifies the critical \texttt{success\_experiences} for completing a subtask or analyzes the \texttt{first\_failure\_subtask\_id} to generate targeted error diagnosis.}
    \label{fig:exp_extract_prompt}
\end{figure*}

\paragraph{Subtask-Level Experience Extraction.}
The model first generates experience for each subtask by analyzing the trajectory. For each completed subtask, the model abstracts detailed success plans. A critical constraint imposed is \textit{Entity Preservation}: specific details (e.g., ``call 1234'') are preserved to ensure subsequent experience abstraction can accurately map them to template variables. For failed subtasks (derived from the first uncompleted subtask id), the model gives analysis about the error root cause and targeted correction guidelines. We provide the used prompt in Figure~\ref{fig:exp_extract_prompt}.

\begin{figure*}[t]
    \centering
    \includegraphics[width=0.85\linewidth]{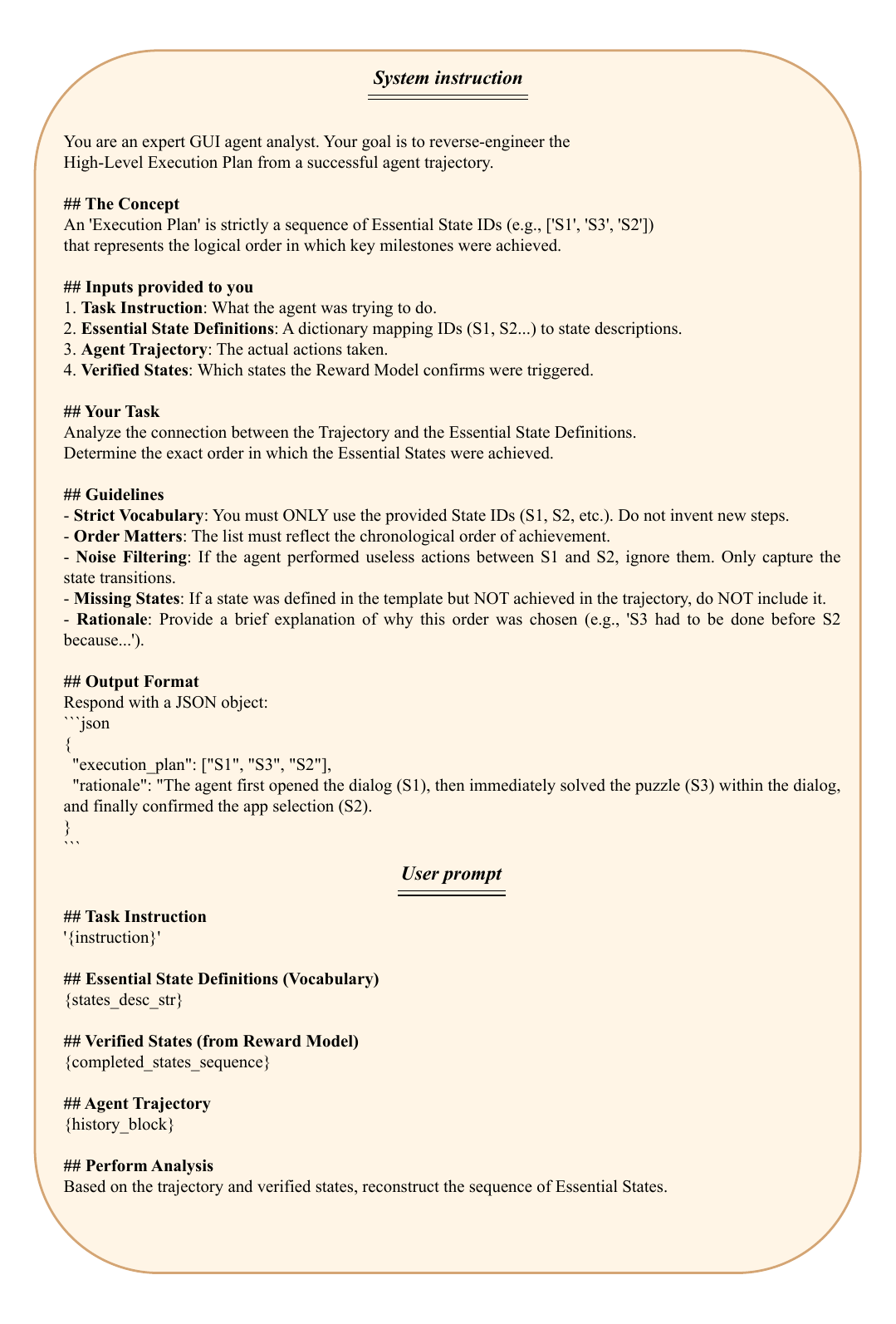}
    \caption{Prompt design for High-Level Workflow Extraction. The system extracts a concrete execution plan from successful traces.}
    \label{fig:exp_extract_prompt2}
\end{figure*}

\paragraph{High-level Workflow Extraction.}
For successful trajectories, we additionally extract the high-level workflow which captures global planning. This process maps the raw agent trajectory to a sequence of subtasks. We provide the used prompt in Figure~\ref{fig:exp_extract_prompt2}.

\section{Experience Abstraction Details}
\begin{figure*}[t]
    \centering
    \includegraphics[width=\linewidth]{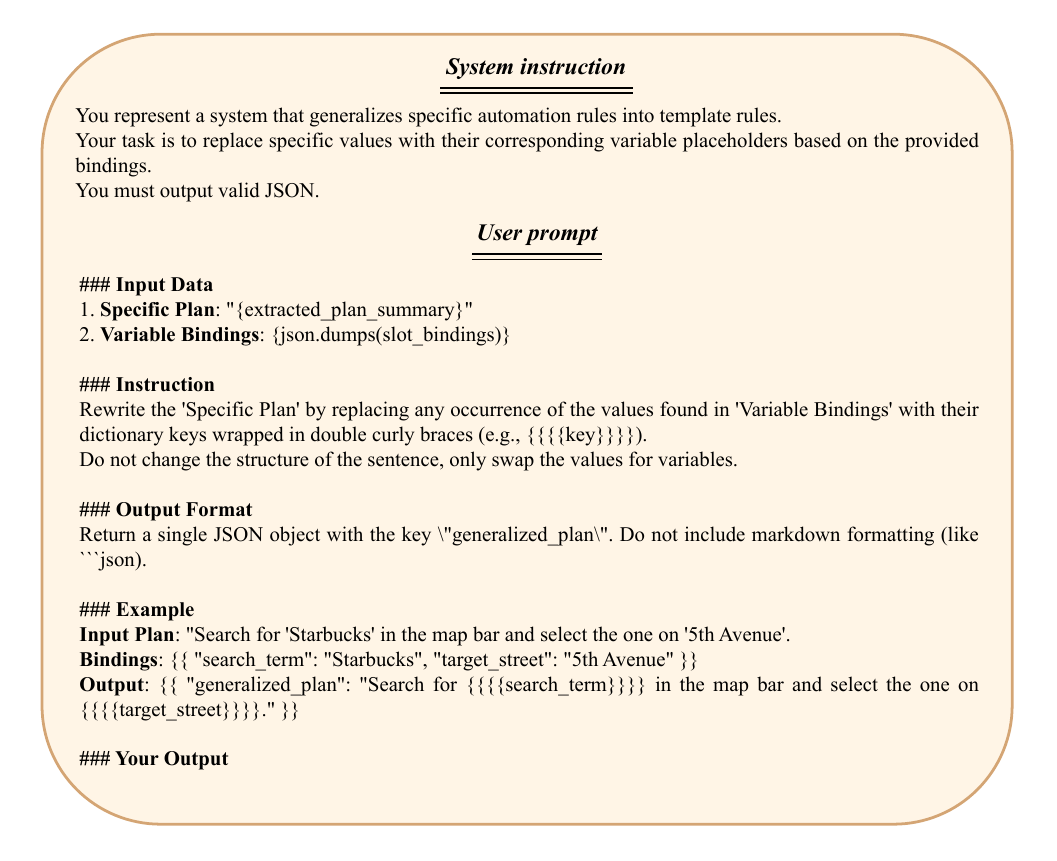}
    \caption{Prompt design for Experience Abstraction. This prompt aims to map concrete entities (e.g., specific cities) to template slots without modifying the original semantic logic.}
    \label{fig:exp_abstract_prompt}
\end{figure*}

\paragraph{Experience Parameterization.}
To transition from task-specific experience to generalizable knowledge, we implement an Experience Parameterization module using \textbf{DeepSeek-V3}. This module transforms obtained concrete experiences into abstracted experiences containing variables, enabling the transfer of learned strategies to new tasks. The input prompt is presented in Figure~\ref{fig:exp_abstract_prompt}.

\paragraph{Experience Ranking Mechanism.}
To facilitate the selection of the most effective abstract knowledge, we implement a scoring strategy. For workflows or subtask skills, we employ an Upper Confidence Bound (UCB) algorithm: $Score = \frac{N_{success}}{N_{total}} + c \sqrt{\frac{\ln N_{global}}{N_{used}+1}}$. This balances the \textit{exploitation} of high-success strategies with the \textit{exploration} of newer, less-tested plans. For \textit{Failure Diagnoses}, we employ a time-decay function $R(t) = \frac{1}{1 + \lambda \Delta t}$ derived from its last updated time. This ensures the agent prioritizes recent error, which is essential for adapting to dynamic UI updates and policy evolving.

\section{The Memory Retrieval Algorithm}

As presented in \textbf{Algorithm \ref{alg:memory_retrieval}}, our memory retrieval procedure consists of serveral hierarchical phases:

\textbf{Task Matching and Variable Extraction.} We first encode the raw instruction into semantic embeddings, using an embedding model Qwen3-Embedding-8B~\cite{qwen3embedding}. We use the embedding model to select top-k similar task templates, which are further processed by an LLM (DeepSeek-V3) to determine the best match template and extract the task-specific variables $\mathcal{V}$.

\textbf{High-Level Workflow Retrieval and Instantiation.} Upon identifying the template, the system retrieves a corresponding abstract workflow from High-level Memory. Since the raw experience is parameterized, the system instantiates it by injecting the variable set $\mathcal{V}$ into predefined placeholders, transforming the workflow into a context-specific, executable plan.

\textbf{Mid-Level Experience Retrieval.} After that, the instantiated plan is enriched with subtask-level guidance and failure correction guidelines. For each pending subtask, the retrieval module queries the memory pool using semantic similarity, similar to process of high-level task templates match. This step aggregates historical success plans and failure diagnoses into the final structured guidance.

\begin{algorithm*}[t]
\centering
\caption{Hierarchical Experience Retrieval \& Instantiation}
\label{alg:memory_retrieval}

\begin{minipage}{0.9\linewidth} 
\small

\begin{algorithmic}[1]
    \Require Current instruction $I$, Task Meta-info $\mathcal{T}_{task}$, Subtask Meta-info $\mathcal{T}_{sub}$, High-level Memory $\mathcal{M}_{high}$, Mid-level Memory $\mathcal{M}_{mid}$, Embedding Model $E(\cdot)$.
    \Ensure Structured Guidance $G$ (Plan + Tips/Warnings).
    
    \AlgPhase{Phase 1: Task Matching \& Variable Extraction}
    \State $e_I \leftarrow E(I)$
    \State Retrieve top-$k$ candidates $\mathcal{C} \subset \mathcal{T}_{task}$ via $\textsc{CosineSim}(e_I, E(\tau))$ for $\tau \in \mathcal{T}_{task}$
    \State $match, \tau^*, V \leftarrow \textsc{LLM\_DecideMatch}(I, \mathcal{C})$ \Comment{Return matched template $\tau^*$ \& variables $V$}
    
    \If{\textbf{not} $match$}
        \State $\tau^* \leftarrow \arg\max_{\tau \in \mathcal{C}} \textsc{Score}(\tau)$ \Comment{Fallback to best analogy}
        \State $V \leftarrow \textsc{ExtractVars}(I, \tau^*)$
    \EndIf
    
    \AlgPhase{Phase 2: Workflow Retrieval}
    \State $P_{raw} \leftarrow \textsc{GetBestPlan}(\mathcal{M}_{high}, \tau^*.id)$ \Comment{Retrieve execution plan with highest success rate}
    \State $P_{filled} \leftarrow \emptyset$
    \For{each subtask step $s_{raw} \in P_{raw}$}
        \State $s_{filled} \leftarrow \textsc{Instantiate}(s_{raw}, V)$ \Comment{Fill placeholders (e.g., \{\{app\}\} $\to$ ``Gmail'')}
        \State $P_{filled}.\text{append}(s_{filled})$
    \EndFor
    
    \AlgPhase{Phase 3: Subtask-Level Experience Retrieval}
    \State $G \leftarrow \text{InitializeGuidance}(P_{filled})$
    \For{each subtask $s \in P_{filled}$}
        \If{$s.label \in \mathcal{M}_{mid}$}
            \State $E \leftarrow \mathcal{M}_{mid}[s.label]$
        \Else
            \State $E \leftarrow \text{best match via } \textsc{CosSim}(E(s.content), \mathcal{M}_{mid}.keys)$
        \EndIf
        \State $E_{tips} \leftarrow \textsc{UCB\_Rank}(E.\text{plan\_summary})$ \Comment{Prioritize using success rate + exploration}
        \State $E_{warn} \leftarrow \textsc{TimeDecay\_Rank}(E.\text{failure\_diagnosis})$ \Comment{Prioritize recent failures}
        \State $G.\text{append}(\{s, E_{tips}, E_{warn}\})$
    \EndFor
    
    \State \textbf{return} $G$
\end{algorithmic}
\end{minipage}
\end{algorithm*}

\section{The Memory Update Algorithm}

The memory update mechanism (\cref{fig:memory_update}) is the core of our self-evolving loop, deriving structured, long-term knowledge from raw trajectories. As demonstrated in \textbf{Algorithm \ref{alg:memory_update}}, this process ensures that the agent learns effectively from past successful and failed trajectories. The procedure consists of the following phases:

\textbf{Global Statistic Update.} The system updates task difficulty metrics via an Exponential Moving Average (EMA) of task success rate. This mechanism dynamically tracks the agent's competence for each task.

\textbf{Mid-Level Experience Abstraction and De-duplication.} The system abstracts subtask feedback by replacing specific parameters with generic placeholders, followed by semantic deduplication. This prevents overfitting to instance-specific values (e.g., filenames) and minimizes the memory redundancy, ensuring a compact and high-quality knowledge base.

\textbf{High-Level Workflow Consolidation.} Successful trajectories are compressed into workflows and merged into the high-level workflow library. The statistical information like success counts are also updated, ensuring that the system effectively prioritizes robust workflows and discards unstable strategies over time.

\begin{algorithm*}[t]
\centering
\caption{Self-Evolving Experience Abstraction \& Update}
\label{alg:memory_update}
\begin{minipage}{0.9\linewidth}
\small
\begin{algorithmic}[1]
    \Require Instruction $I$, Trace $\mathcal{T} = \{(S_i, \text{bindings}_i)\}$, Feedback $\mathcal{F}$ (Success/Correction), Global Rate $SR$.
    
    \State \textbf{Input:} Executed trace with instantiated templates and variable bindings.

    \AlgPhase{Phase 1: Global Statistic Update (EMA)}
    \State $SR(I) \leftarrow \gamma \cdot SR(I) + (1-\gamma) \cdot \text{Score}(\mathcal{T})$ \Comment{Update moving avg success rate}

    \AlgPhase{Phase 2: Mid-Level Memory Update (Subtask Experience)}
    \For{each subtask $S_i \in \mathcal{T}$ containing feedback in $\mathcal{F}$}
        \State $key \leftarrow (S_i.\text{pkg}, S_i.\text{label})$
        \State $exp_{raw} \leftarrow \mathcal{F}.\text{get\_content}(S_i)$ \Comment{Get summary if success, diagnosis if fail}
        
        \State $exp_{abs} \leftarrow \textsc{LLM}(exp_{raw}, S_i.\text{bindings})$ \Comment{Parameterize: replace values with slots} 
        
        \State $Target \leftarrow \mathcal{M}_{mid}[key].\text{select\_list}(S_i.\text{status})$ \Comment{Select PlanSummary or FailureDiagnosis list}
        \State $sim \leftarrow \max_{e \in Target} \textsc{CosSim}(exp_{abs}, e)$ \Comment{Check for semantic redundancy}
        
        \If{$sim \ge \delta_{dup}$}
            \State Update counts ($N_{success}, N_{used}$) \Comment{Update metrics for UCB/Decay}
            \If{new experience is more detailed}
                \State Update content \Comment{Keep higher quality experience}
            \EndIf
        \Else
            \State $Target.\text{append}(exp_{abs})$ \Comment{Register novel subtask experience}
        \EndIf
    \EndFor

    \AlgPhase{Phase 3: High-Level Memory Update (Task Planning)}
    \If{Task is Success}
        \State $Seq \leftarrow [S_1.id, \dots, S_T.id]$ \Comment{Extract abstract execution flow}
        \State $\mathcal{P} \leftarrow \mathcal{M}_{high}[I].\text{plans}$
        
        \If{$Seq \in \mathcal{P}$} \Comment{Exact workflow matching}
            \State Update success count \& avg steps for $Seq$
            \State Update rationale if new one is better
        \Else
            \State $\mathcal{P}.\text{add}(\{Seq, \text{count}:1\})$ \Comment{Record new successful workflow}
        \EndIf
        \State Sort $\mathcal{P}$ by UCB algorithm.
    \EndIf

\end{algorithmic}
\end{minipage}
\end{algorithm*}

\section{Qualitative Examples}

To qualitatively evaluate the effectiveness of the proposed UI-Mem framework, we present visualized examples illustrating successful long-horizon planning, error correction via failure diagnosis, and remaining challenges in visual grounding.

\begin{figure*}[t]
    \centering
    \includegraphics[width=\linewidth]{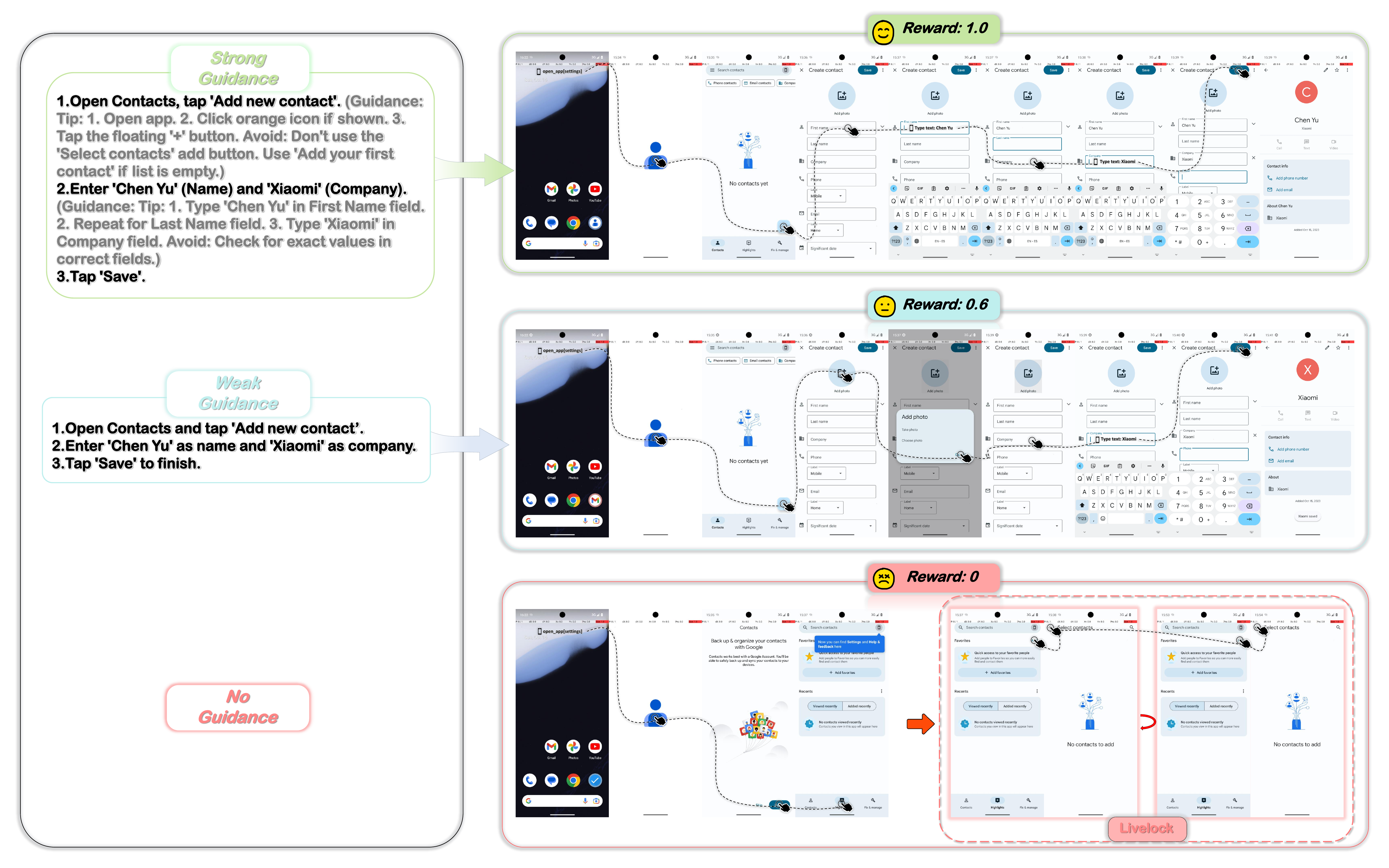} 
    \caption{Trajectory analysis on the task ``Create a new contact for Chen Yu, with the organization listed as Xiaomi.'' \textbf{Top (Reward 1.0):} Full memory guidance yields a perfect execution. \textbf{Middle (Reward 0.6):} Incomplete execution where the agent correctly fills the organization but misses the name. \textbf{Bottom (Reward 0):} Lack of memory results in a total failure.}
    \label{fig:contacts_rollout}
\end{figure*}

\paragraph{Impact of Memory Guidance.} 
We demonstrates how different strengths of guidance contribute to the progress of task completion. We provide visualizations of the agent's rollout trajectories for the specific task instruction \textit{``Create a new contact for Chen Yu, with the organization listed as Xiaomi.''} As shown in the top row of Figure~\ref{fig:contacts_rollout}, full memory guidance leads to \textbf{High Task Progress (Reward 1.0)}, where the agent executes a flawless trajectory, mapping ``Chen Yu'' to the Name field and ``Xiaomi'' to the Company field before saving. In contrast, the middle row illustrates \textbf{Partial Completion (Reward 0.6)} under weak guidance (only workflow). Here, the agent exhibits incomplete task execution: while it correctly inputs the organization ``Xiaomi,'' it fails to input the name ``Chen Yu'' and deviates into irrelevant actions (e.g., adding a photo), resulting in a contact saved with missing information. Finally, without any prior guidance, the agent fails to achieve any progress in completing the task, repeatedly toggling between the search bar and the empty list view. This comparison validates our design that injecting different levels of guidance increases the outcome diversity within a rollout group given the same task instruction.

\begin{figure*}[t]
    \centering
    \includegraphics[width=\linewidth]{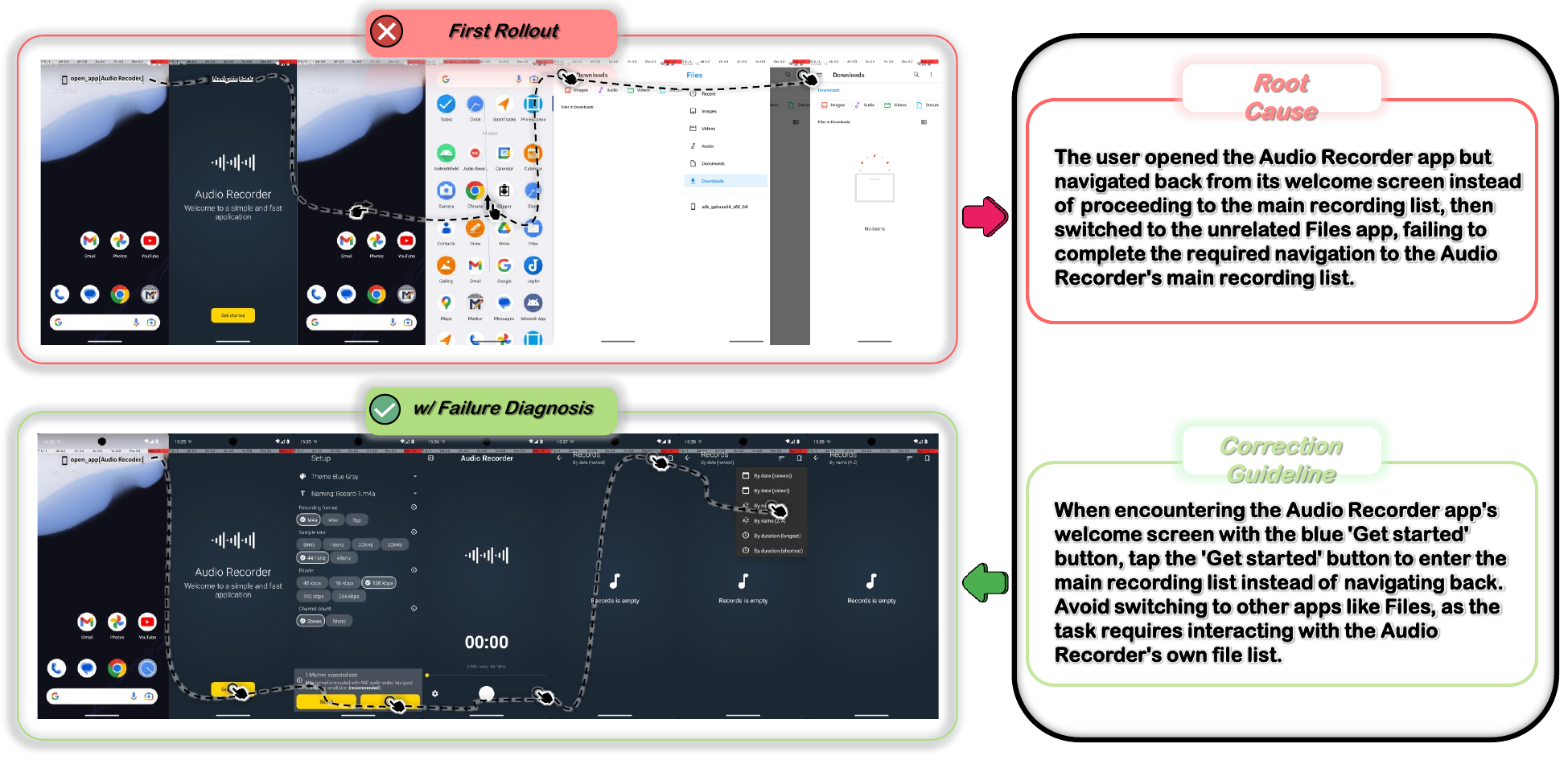} 
    \caption{Visualizing the Failure Diagnosis mechanism on the task ``Enter the file list and sort the files by name (A-Z).'' The system identifies the navigation error in the first rollout and generates a specific \textbf{Correction Guideline}, enabling success in the second-round attempt.}
    \label{fig:error_correction}
\end{figure*}

\paragraph{Error Correction via Diagnosis.} 
We demonstrate the effectiveness of integrating failure pattern into our memory using another Audio Recorder task: \textit{``Enter the list and sort the files by name (A-Z).''} Initially, as shown in the top of Figure~\ref{fig:error_correction} (First Rollout), the agent fails by misinterpreting the welcome screen's navigation, erroneously exiting the app to an unrelated ``Files'' application. After this rollout, our memory incorporates the \textbf{Failure Diagnosis} identifying the root cause: \textit{``The user opened the Audio Recorder but navigated back... instead of proceeding to the main recording list.''} Equipped with this insight, during the second attempt (\textbf{w/ Failure Diagnosis}), the agent retrieves a \textbf{Correction Guideline} instructing it to \textit{``Tap the `Get started' button... avoid switching to other apps.''} Given this guidance, the agent correctly enters the main list view and successfully performs the sorting operation.

\begin{figure*}[t]
    \centering
    \includegraphics[width=\linewidth]{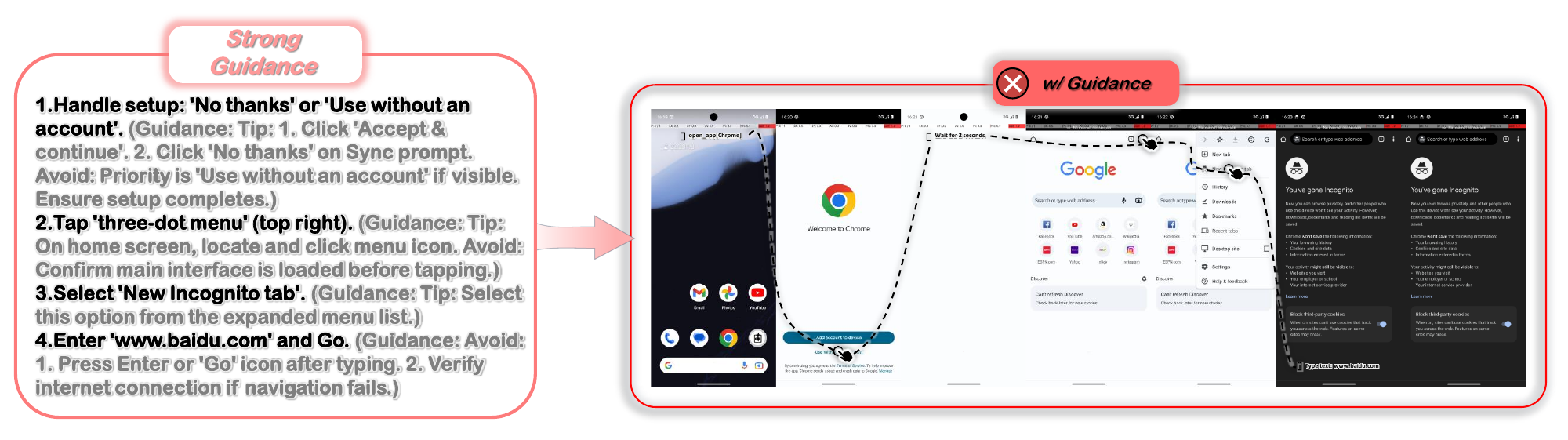} 
    \caption{A failure case in the Browser task. Despite fine-grained guidance (opening Incognito), the agent fails to visit the URL due to a specific visual grounding error during the final action step.}
    \label{fig:failure_case}
\end{figure*}

\paragraph{Plan-Execution Gap in Failure Cases.} 
Despite the overall effectiveness of our memory guidance, there are remaining failure modes due to unexpected grounding errors. In Figure~\ref{fig:failure_case}, the agent is required to complete the task \textit{``Create a new incognito tab and visit www.baidu.com.''} The agent retrieves a fine-grained guidance to handle the distinctive ``Welcome'' popup and navigates the Chrome menu to select ``New Incognito tab.'' However, the rollout fails at the final step due to a visual grounding error. Although the planner intends to visit the URL, the agent fails to correctly select the address bar (Omnibox) to type ``www.baidu.com.''

\end{document}